\documentclass{article}

\PassOptionsToPackage{numbers,sort&compress}{natbib}
\usepackage[preprint]{neurips_2026}

\usepackage[utf8]{inputenc}
\usepackage[T1]{fontenc}
\usepackage{microtype}
\usepackage{hyperref}
\usepackage{url}
\usepackage{booktabs}
\usepackage{amsmath}
\usepackage{amsfonts}
\usepackage{graphicx}
\usepackage{caption}
\usepackage{subcaption}
\usepackage{placeins}
\usepackage{threeparttable}
\usepackage{multirow}
\usepackage{float}
\usepackage{enumitem}
\usepackage{tabularx}
\usepackage[most]{tcolorbox}
\usepackage{xcolor}
\usepackage{bbm}
\usepackage{pifont}
\usepackage{caption}
\usepackage{wrapfig}
\usepackage[table]{xcolor}
\usepackage{bold-extra}

\newcommand{\cmark}{\ding{51}}
\newcommand{\xmark}{\ding{55}}

\graphicspath{{figures/}}

\definecolor{darkblue}{rgb}{0, 0, 0.5}
\hypersetup{colorlinks=true, citecolor=darkblue, linkcolor=darkblue, urlcolor=darkblue}

\tcbset{
  aibox/.style={
    width=\linewidth,
    top=8pt,
    bottom=4pt,
    colback=blue!6!white,
    colframe=black,
    colbacktitle=black,
    enhanced,
    center,
    attach boxed title to top left={yshift=-0.1in,xshift=0.15in},
    boxed title style={boxrule=0pt,colframe=white},
  }
}
\newtcolorbox{AIbox}[2][]{aibox,title=#2,#1}

\tcbset{
  dcbox/.style={
    width=\linewidth,
    top=8pt,
    bottom=4pt,
    colback=red!6!white,
    colframe=black,
    colbacktitle=black,
    enhanced,
    center,
    attach boxed title to top left={yshift=-0.1in,xshift=0.15in},
    boxed title style={boxrule=0pt,colframe=white},
  }
}
\newtcolorbox{DCbox}[2][]{dcbox,title=#2,#1}

\tcbuselibrary{breakable}

\newcommand{\sizhe}[1]{{\color{purple} [Sizhe]: #1}}
\newcommand{\sewon}[1]{{\color{violet} [Sewon]: #1}}

\newcommand{\tightparagraph}[1]{\vspace{-.7em}\paragraph{#1}}

\newcommand{\myskip}[1]{}

\title{Reference-Based Distillation Detection in LLMs}

\author{
  \textbf{Rajat Rawat}$^{1}$ \quad
  \textbf{Sizhe Chen}$^{1}$ \quad
  \textbf{Akshay Anand}$^{1}$ \quad
  \textbf{Michael Duan}$^{2}$ \\
  \textbf{Bob Rotsted}$^{3}$ \quad
  \textbf{Sewon Min}$^{1}$ \\[0.6em]
  $^{1}$University of California, Berkeley \quad
  $^{2}$University of Southern California \quad
  $^{3}$OpenAI %
  \\[0.4em]
  \texttt{rajat\_rawat@berkeley.edu}
}

\begin{document}

\maketitle

\begin{abstract}
Model distillation---training on outputs from stronger third-party models---is widely used to boost performance, but raises concerns about unfair advantages and policy violations. This motivates a fundamental question: \emph{can we detect whether a model was distilled from another?} We show that, while identifying a teacher model from a student in isolation is highly challenging, it becomes tractable in a \textbf{reference-based setting}: given a model and an earlier-generation checkpoint from the same lineage, we can identify the teacher model used to train the later checkpoint.
We introduce a distillation detection method based on reference-based membership inference. By comparing how strongly a student model preferentially aligns with outputs from different candidate teachers relative to a reference checkpoint, our method identifies the most likely teacher and detects evidence of distillation.
To handle unknown distillation pipelines such as hidden prompts, we infer proxy prompt templates directly from model outputs. We additionally identify a distinctive glyph-level signal specific to o1/o3 models. %
Evaluating distillation detection is challenging because modern model lineages are already heavily entangled. To address this, we develop a hybrid evaluation spanning both controlled distillation experiments and real-world models. Across both settings, our approach recovers the true teacher with near-perfect accuracy in single-teacher distillation scenarios, even when the underlying distillation pipeline is largely unknown. We further introduce statistical tests for both teacher attribution and distillation detection, and extend our framework to open-world settings where no teacher is guaranteed to be present among the candidates. Applying our method to contemporary models yields new evidence regarding potential distillation relationships involving QwQ, DeepSeek-R1, and GPT-OSS.\footnote{Code is available at \url{https://github.com/RajatRawat-creator/DistillDetect}.}
\end{abstract}

\section{Introduction}\label{sec:intro}
Large language models (LLMs) are increasingly trained via model distillation, where a student is fine-tuned on outputs from a stronger teacher. This approach %
has become a standard way to transfer capabilities such as reasoning and instruction following \citep{hinton2015distilling,wang2022selfinstruct,toshniwal2024openmathinstruct2,muennighoff2025s1}. While effective and widely adopted, distillation creates ecosystem-level risks: a competitor can query a proprietary model at scale and train a competing system on its outputs, cheaply acquiring capabilities that would otherwise require new architectures or datasets. This raises concerns about unfair advantage and policy violations. Major providers (e.g., Google, OpenAI, Anthropic) have responded with system-level defenses based on request metadata \citep{googledistillation,openaidistillation,anthropicdistillation}, but these require usage logs and infrastructure access and do not address detection from model behavior.

This motivates a fundamental problem in model auditing: \emph{given a model, can we determine whether a teacher was used to train it?} Answering this is important for intellectual property protection, accountability, and transparency when training details are undisclosed, but is challenging due to both the difficulty of membership inference in modern LLMs and the opacity of distillation pipelines, including unknown prompts, data sources, and filtering. Detecting distillation in this absolute sense is quite difficult; however, we find that a reference-based setting is more tractable: \emph{given a model and an earlier-generation checkpoint of itself, can we determine whether a teacher was used to train the later checkpoint from the earlier one, and if so, which teacher it was?}

\begin{figure}[t]
    \centering
    \includegraphics[width=\columnwidth]{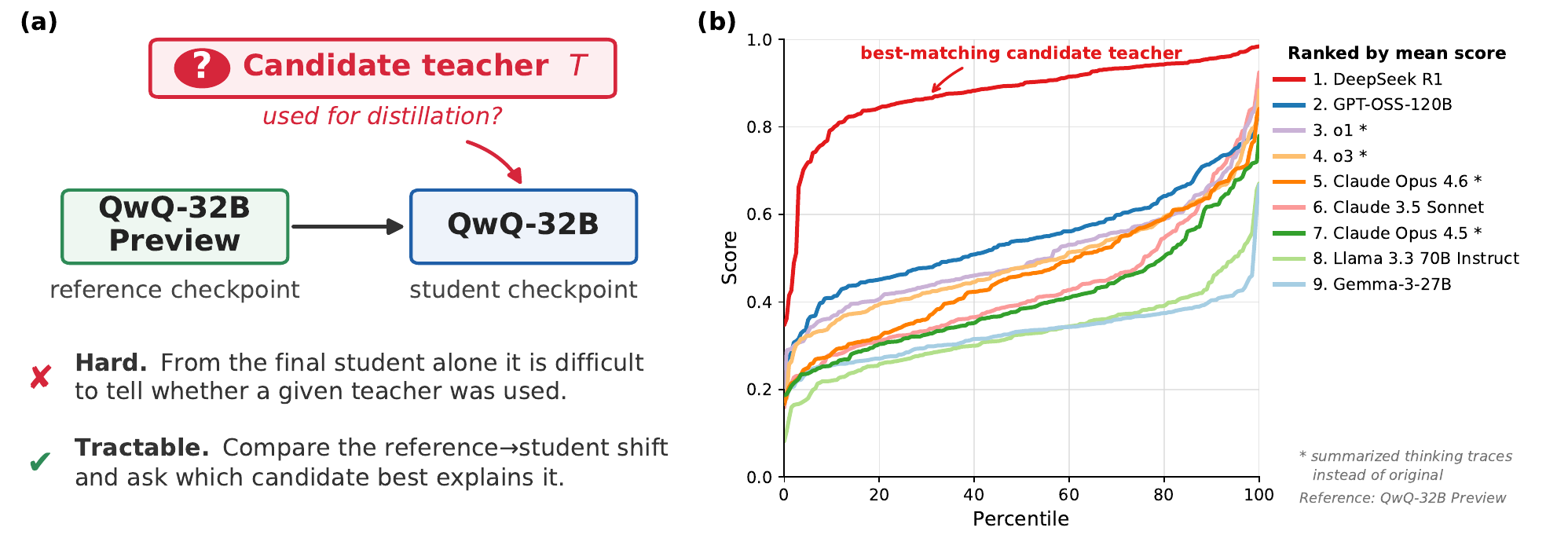}
    \vspace{-1.2em} \caption{Reference-based distillation detection. (a) Rather than attributing distillation from the final student alone, we measure the shift relative to a reference model (e.g., an earlier model from the same family). (b) Scoring candidate teachers for QwQ-32B identifies DeepSeek R1 (red) as the best match.}
    \label{fig:distillation}
\end{figure}

In this work, we propose a teacher identification method based on contrastive, reference-normalized likelihoods (\S\ref{sec:method}; Figure~\ref{fig:distillation}(a)). Given candidate teachers, we score each by how strongly the student prefers that teacher's outputs relative to a reference model, and select the teacher whose signal consistently dominates.
The key insight is that likelihoods taken relative to the earlier checkpoint---not absolute ones---carry the signal: standard membership inference metrics (e.g., raw likelihood, ZLIB-normalized likelihood, Min-$K$ probability) fail to reliably recover the true teacher, whereas our reference-based formulation remains robust even under opaque and heterogeneous pipelines. The method requires only black-box access, extends naturally to hidden-prompt settings via proxy prompt inference, and is paired with a per-probe significance test for teacher attribution. %

Beyond likelihood-based attribution, we identify a distinctive glyph-level signature of o1/o3 models: unusually frequent use of non-ASCII Unicode characters. We exploit this signal through a reference-normalized loss gap between ASCII and Unicode serializations, yielding a complementary indicator of o1/o3 distillation that persists in distilled students. %

Evaluating distillation detection is itself challenging because model lineage is already deeply entangled across modern LLMs. We therefore design a hybrid evaluation method combining controlled experiments and real-world models (\S\ref{sec:teacher-identification}), focusing on a canonical setting: single-teacher, last-stage, SFT-style distillation. Controlled experiments use student--teacher pairs known not to be distilled from one another, ensuring clean ground truth at the cost of a more synthetic setup; real-world models better reflect practical conditions but introduce ambiguity, as candidate teachers may themselves be distilled from each other. Across both settings, our method reliably recovers the true teacher with near-perfect accuracy, significantly outperforming baselines, and remains robust under domain shift between distillation and detection data. To separate genuine distillation from coincidental similarity, we pair the detection threshold with a significance test on the per-probe margin and require agreement across two separate prompt sets. Cross-validation results suggest that the resulting threshold generalizes well to previously unseen student models. %

Finally, we extend our framework to the open-world setting, where no teacher is guaranteed to exist among the candidates tested, and apply it to contemporary models to examine open provenance questions (\S\ref{sec:openquestions}; Figure~\ref{fig:distillation}(b)). Through case studies on QwQ, DeepSeek-R1, and GPT-OSS, we quantify empirical evidence of potential distillation and identify likely teacher sources when such signals exist.

Together, these findings establish the feasibility of model-level auditing for distillation in modern LLMs, advancing transparency and accountability in increasingly opaque model development pipelines.
While our study focuses on a simplified setting, we hope this work motivates future research on attribution under more complex regimes, including multi-teacher and multi-stage distillation.

\section{Related Work}\label{sec:related}
\paragraph{Distillation in LLMs}

Model distillation was introduced for compressing knowledge into smaller models in image models \citep{hinton2015distilling}.
Recent work has adopted distillation to obtain strong LLMs \citep{kim2016sequence,sanh2019distilbert,jiao2019tinybert}, such as Phi \citep{abdin2024phi}, Alpaca \citep{taori2023alpaca}, etc. The performance gain from distillation is improved when the teacher is a reasoning LLM %
\citep{muennighoff2025s1}. %

While effective and widely adopted, distillation creates ecosystem-level risks: model developers may query competitor models at scale and train systems on their outputs to replicate capabilities at a fraction of the original development cost. This has emerged as an increasingly imminent concern, with industry-scale distillation repeatedly reported by major AI providers~\citep{googledistillation,anthropicdistillation,openaidistillation}.

\tightparagraph{Distillation detection in LLMs}

The growing use of distillation to reproduce competitor capabilities has consequently prompted major AI providers to deploy detection systems.
Existing industry detection primarily relies on request-side monitoring and infrastructure signals rather than model behavior itself, such as request metadata, IP address correlation, and infrastructure-level indicators~\citep{anthropicdistillation}. 

Another line of work studies watermark-based detection, where the auditor controls the teacher model and embeds detectable signals into generated outputs~\citep{sander2024watermarkingmakeslanguagemodels}. However, these approaches require strong assumptions about provider-side access, and recent work shows that such watermarks can often be weakened or removed during downstream training~\citep{pan2025llmwatermarksrobustlyprevent}.

\citep{zhang2025detectingdistillationdatareasoning} studies a different but related problem: detecting which prompts are used for generating continuations used in a model's distillation. %

In summary, distillation detection has attracted growing attention in both industry and academia. These approaches target absolute, provider-level detection. To the best of our knowledge, there is no prior work on reference-based LLM detection from model behavior alone, and we present this work as a first step in that direction.

\tightparagraph{Distillation detection in non-LLM settings}
Distillation detection has also been studied outside the LLM setting, particularly for image models. Prior work detects distilled image classifiers and generative models using synthesized inputs and statistical scoring~\citep{shi2025knowledge}, prompt-level fingerprints in text-to-image models~\citep{zhang2024trainingdataattributionmodel}, and shadow-model-based provenance verification comparing generalization behavior against suspicious models~\citep{xie2025trainingdataprovenanceverification}.

\tightparagraph{Membership Inference}

Our approach relies on membership inference, which determines whether a particular example was used during model training \citep{shokri2017membership} according to %
the intuition that training examples receive lower loss than non-training examples. It has been shown that 
comparing against a reference output distribution yields stronger membership inference signal \citep{carlini2022lira,watson2022importancedifficultycalibrationmembership}. A recent work extends membership inference to decide whether a model is trained on a dataset \citep{maini2024llmdatasetinferencedid}.
We build on these literature but study a different problem: whether a model is trained on another model's generations.

\section{Method}\label{sec:method}
\paragraph{LLM Distillation}
We consider supervised fine-tuning (SFT)-based distillation, a common approach for transferring behavior from a teacher LLM to a smaller student. Let $S_0$ be a base student model and $T$ a teacher model. Given inputs $\mathcal{X}=\{x_i\}_{i=1}^N$, where each $x_i$ includes the prompt template used to query the teacher, the teacher produces outputs $y_i=T(x_i)$, yielding the distillation dataset $\mathcal{D}_T=\{(x_i,y_i)\}_{i=1}^N$. Fine-tuning $S_0$ on $\mathcal{D}_T$ produces the distilled student $S$.

\tightparagraph{Distillation detection}
\label{subsec:method-distillation-detection}
Given a subject model $S$ and candidate teachers $\mathcal{T}=\{T_1,\dots,T_K\}$, our goal is to determine whether $S$ was distilled from any teacher in $\mathcal{T}$, and if so, identify the teacher. As an initial step, we assume $S$ was distilled from at most one teacher in $\mathcal{T}$. We assume the detector has logits access to $S$ and $\mathcal{T}$, but not to the prompts or templates used for distillation; instead, the detector queries $S$ using its default chat template and public prompts, testing whether distillation signals remain detectable under this limited-information setting.

A natural first attempt is to detect distillation in an \emph{absolute} sense: scoring candidate teacher outputs directly under $S$---for instance, via raw likelihood or standard membership-inference metrics---without any point of comparison. We find this unreliable. We instead adopt a \emph{reference-based} formulation: rather than asking whether $S$ explains a teacher's outputs in absolute terms, we ask whether it explains them \emph{better than} an earlier-generation reference checkpoint $R$ does. The rest of this section develops this reference-based approach for teacher identification (\S\ref{subsec:teacher-identification-method}) and the binary detection problem (\S\ref{subsec:distillation-detection-method}).

\subsection{Teacher Identification}\label{subsec:teacher-identification-method}

We first consider the simpler setting of \textbf{teacher identification}, assuming a single true teacher exists in $\mathcal{T}$, as also studied in prior computer vision work~\citep{shi2025knowledge}. Formally, the task maps a student $S$ and candidate set $\{T_1, \ldots, T_K\}$ to the index of the true teacher: $(S, \{T_1, \ldots, T_K\}) \mapsto y \in [K].$
Our approach proceeds as follows:
\begin{enumerate}[leftmargin=17pt, topsep=0pt,itemsep=1pt]
    \item Construct proxy inputs $\hat{\mathcal{X}}$ from open-source data, which may differ from the original inputs $\mathcal{X}$.
    \item Query each teacher $T_k$ to obtain outputs for every $x_i \in \hat{\mathcal{X}}$.
    \item Use a scoring function $\mathbf{f}$ to compute alignment $\mathbf{f}(T_k(x_i), S)$ between each teacher output and $S$.
\end{enumerate}
We identify the teacher using two complementary decision rules: (i) selecting the teacher with the highest mean alignment score across $\hat{\mathcal{X}}$, and (ii) sorting per-teacher scores over $\hat{\mathcal{X}}$ and selecting the teacher that wins the largest fraction of rank-matched comparisons. The latter captures distributional preference without requiring per-prompt dominance. We report \textbf{Per-sample} as the fraction of rank-matched comparisons won by the true teacher, and \textbf{Agg.} as the better of the two decision rules. 

\tightparagraph{How to choose \textbf{$\mathbf{f}$}?}
Concretely, the non reference-based approach instantiates $\mathbf{f}$ as a direct comparison between $S(x)$ and $T_k(x)$---via lexical or semantic similarity ($n$-gram overlap, embedding metrics), or by scoring $T_k(x)$ by its log-likelihood under $S$. We find these unreliable: log-likelihood, in particular, can favor teacher outputs that are easy to model in general rather than specifically aligned with $S$ (empirical evidence in Table~\ref{tab:teacher_identification_controlled}).

This motivates the reference-based formulation. Following reference-based membership inference~\citep{watson2022importancedifficultycalibrationmembership}, we calibrate the log-likelihood under $S$ using an earlier generation checkpoint reference model $R$, which was ideally not exposed to any teacher candidate, or otherwise less affected by the distillation. For example, $R$ may be an earlier checkpoint in the same family (e.g., Llama 3.1 Instruct as $R$ for Llama 3.2 Instruct), a prior training stage (e.g., a base model when $S$ is instruction-tuned), or a smaller variant (e.g., GPT-OSS 20B as $R$ for GPT-OSS 120B). We evaluate several choices of $R$ (\S\ref{subsec:abl_ref_model}) and find that a suitable reference can typically be identified in realistic auditing scenarios.

For each example, we compute the normalized alignment score
\[
\mathbf{f}(T_k(x_i), S) = -\big(\ell_S(x_i, y_i^{(k)}) - \ell_R(x_i, y_i^{(k)})\big),
\]
where $\ell_M(x, y)$ is the average token-level negative log-likelihood of $y$ given $x$ under model $M$. This measures how much better the student explains a candidate teacher's output relative to the reference. We aggregate these scores per teacher using the mean-score and sorted per-sample rules above.

\subsection{Distillation Detection}
\label{subsec:distillation-detection-method}
Beyond identifying a single teacher, we consider the binary \emph{distillation detection} problem: given $S$ and candidate set $\{T_1, \ldots, T_K\}$, predict $y \in \{0,1\}$ indicating whether $S$ was distilled from any teacher in the set.\footnote{We assume $S$ is distilled from at most one candidate teacher.}
Using the reference-normalized scores $\mathbf{f}$ from \S\ref{subsec:teacher-identification-method}, we define the \textbf{margin} $\delta = \bar{s}_1 - \bar{s}_2$, where $\bar{s}_k = \frac{1}{N}\sum_{i=1}^N \mathbf{f}(T_k(x_i), S)$ and $\bar{s}_1 \geq \bar{s}_2$ are the top two mean scores among candidates. If the true teacher is present, it should dominate the pool, yielding a large $\delta$; if absent, scores are more uniform and $\delta$ stays small.
Given a calibration set, we select a threshold $\tau$ by trying each observed training margin as a cutpoint and choosing the value that maximizes F1 on the calibration examples. Crucially, $\tau$ is selected using only the training examples and never the held-out margins. For each held-out evaluation cell, its margin is compared against the corresponding fold-specific threshold. The final model-level detection additionally requires the margin to significantly exceed $\tau$ on both prompt sets and for both prompt sets to identify the same teacher, as defined in \S\ref{subsec:significance}.
\subsection{Statistical Significance}
\label{subsec:significance}
Both tasks above output a point estimate---a teacher ranking for identification
and a margin $\delta$ for detection---so for each we add a test that checks whether
the signal is real rather than noise. We start with identification, where the true
teacher is known.

\tightparagraph{Per-probe top-1 consistency (identification).}
With the true teacher $t^\star$ known in advance, we ask the most direct question: on
how many \emph{individual} probes is $t^\star$ the top-scoring candidate? We count a
win on probe $x_i$ when $t^\star$ outscores every other candidate,
\[
X_i = \mathbbm{1}\!\left[\arg\max_{k}\,\mathbf{f}(T_k(x_i), S) = t^\star\right],
\]
and report the win rate $\hat{\rho} = \tfrac{1}{N}\sum_{i=1}^N X_i$; a \emph{majority} means
$\hat{\rho} > 0.5$. To rule out chance, note that if ranking were random $t^\star$ would top a probe with probability $1/K$, making the $N$ wins Binomial; we test $H_0:\rho \le 1/K$ against $H_1:\rho > 1/K$ with a one-sided exact binomial test. Because the test is exact, it needs no normality check or Wilcoxon fallback, and because $t^\star$ is fixed in advance (not picked as the best of $K$), it needs no Bonferroni correction---which is why this is an identification test, complementing the detection test below.

\tightparagraph{Margin significance (detection).}
The margin $\delta$ is a point estimate; to decide whether an observed $\delta$ reflects distillation rather than sampling noise, we attach an inferential test. Let $t_1$ and $t_2$ denote the top- and second-ranked candidates by mean score, and define the paired per-probe difference
\[
d_i = \mathbf{f}(T_{t_1}(x_i), S) - \mathbf{f}(T_{t_2}(x_i), S).
\]
Pairing within a probe cancels per-prompt difficulty shared across candidates, and the sample mean of $d$ recovers $\delta$. We test against the calibrated threshold $\tau^*$ rather than against zero: at $N\!\approx\!200$, even a negligible margin can be significant against zero, including for non-distilled models. Concretely, for $H_0\!:\delta \leq \tau^*$ against $H_1\!:\delta > \tau^*$, we use a one-sided $t$-test with statistic $t = (\delta - \tau^*)/(s_d/\sqrt{N})$ on $N\!-\!1$ degrees of freedom, and report a $95\%$ confidence interval for $\delta$ from $10{,}000$ probe-level bootstrap resamples as an effect-size diagnostic where shown (Table~\ref{tab:o1_significance}). Since $d$ is occasionally non-normal (Shapiro--Wilk), we additionally compute a one-sided Wilcoxon signed-rank test on the shifted differences $d_i - \tau^*$---a distribution-free check of the same hypothesis---and base the decision on this test whenever the normality assumption is rejected (Shapiro $p < 0.05$). The decision uses the margin over the runner-up $t_2$: since $t_2$ is the closest competitor, clearing $\tau^*$ against it would generally imply clearing $\tau^*$ against every weaker candidate $T_k$ ($k \neq t_1$).

Because $t_1$ is selected as the best of $K$ candidates, we Bonferroni-correct the selection by reporting $p^* = \min(1,\, p \cdot (K\!-\!1))$, where $p$ is the $p$-value of the governing test ($t$ or Wilcoxon). We call a target distilled from teacher $T$ only if the corrected test rejects $H_0$ ($p^* < 0.05$) on \emph{both} prompt sets and both name the same teacher; requiring cross-prompt agreement makes the call teacher-specific rather than family-generic.

\section{Experiment: Teacher Identification}\label{sec:teacher-identification}
We first validate our teacher identification method.

\tightparagraph{Disclaimer.}
Our evaluation focuses on a canonical setting---single-teacher, SFT-style final-stage distillation---and isolates this minimal setup; real-world pipelines may involve multiple teachers, multi-stage use, and diverse roles (e.g., generation, rewriting), which we defer to future work.

\subsection{Experimental Setup}\label{subsec:teacher-identification-setup}
Because modern LLM lineages are deeply entangled, ground-truth attribution is often ambiguous. We therefore use a hybrid evaluation: \textbf{controlled experiments}, where we distill models ourselves to isolate key variables, and \textbf{models in the wild}, where real-world distilled models test robustness under unknown prompts, instructions, filtering, and training pipelines.

\tightparagraph{Controlled Experiments}
Since most state-of-the-art LLMs may already have been distilled from each other, we select teacher--student pairs from distinct U.S.-organization model families, where legal constraints limit cross-company distillation \citep{philipp2026prompt}, and additionally include Qwen models to broaden coverage---though we acknowledge that prior distillation is harder to rule out in these cases.

We consider four base student models ($S_0$): Llama-3.2-3B-Instruct~\citep{meta_llama32_3b_instruct}, Gemma-3-4B-PT~\citep{gemma3_technical_report}, Qwen2.5-1.5B, and Qwen2.5-3B~\citep{qwen25_technical_report}. We distill each from three teachers: Qwen3-8B~\citep{qwen3_technical_report}, Llama-3.3-70B-Instruct~\citep{meta_llama33_70b_instruct}, %
and GPT-OSS-120B~\citep{gptoss_model_card}, using prompts from OpenMathInstruct-2 (OMI-2) and s1~\citep{toshniwal2024openmathinstruct2,muennighoff2025s1}, yielding $4 \times 3 \times 2 = 24$ distilled students. We retain the 19 for which distillation improves utility over the base model. At detection time, we additionally include Gemma-3-27B-IT~\citep{gemma3_technical_report} as a candidate teacher, forming a four-way classification task. 

\tightparagraph{Real-World Distilled Models}
We consider (1) six DeepSeek-R1 distill models officially distilled by DeepSeek;\footnote{\url{https://huggingface.co/deepseek-ai/DeepSeek-R1-Distill-Qwen-7B\#deepseek-r1-distill-models}} (2) \texttt{s1.1-32B},\footnote{\url{https://huggingface.co/simplescaling/s1.1-32B}} distilled from R1-generated datasets; and (3) X-Coder-SFT-Qwen3-8B,\footnote{\url{https://huggingface.co/IIGroup/X-Coder-SFT-Qwen3-8B}} a code-focused model trained via SFT on synthetic code data from multiple reasoning models (including \texttt{DeepSeek-R1} and \texttt{Qwen-3-235B-A22B}) with prompts from \texttt{o3-mini}.

For the pool of teacher candidates, we consider 10 models:
Claude Opus 4.5, Claude Opus 4.6, Claude 3.5 Sonnet, DeepSeek R1, GPT-OSS-120B, Gemma-3-27B-it, Llama-3.3-70B-Instruct, QwQ-32B-Preview, o1, and o3---forming a ten-way classification task.

\begin{table}[!t]
\centering
\footnotesize
\begin{tabular}{l rr rr rr}
\toprule
  \multirow{2}{*}{Method}
      & \multicolumn{2}{c}{Llama 3.3 70B Instruct}
      & \multicolumn{2}{c}{Qwen-3-8B}
      & \multicolumn{2}{c}{GPT-OSS-120B} \\
  \cmidrule(lr){2-3} \cmidrule(lr){4-5} \cmidrule(lr){6-7}
  & Agg. & Per-sample
  & Agg. & Per-sample
  & Agg. & Per-sample \\
\midrule
  Random
      & 25.0 & 25.0 & 25.0 & 25.0 & 25.0 & 25.0 \\
  Length Matching Baseline
      & 42.9 & 58.5 & 33.3 & 19.1 & 33.3 & 23.2 \\
\midrule
  \multicolumn{7}{l}{\textbf{\em Non Reference-Based Methods}} \\
  \multicolumn{7}{l}{\quad\textit{String-based}} \\
  \quad Multi n-gram Jaccard
      & 100.0 & 87.6 & 33.3 & 27.3 & 0.0 & 0.5 \\
  \quad Symmetric Coverage
      & 85.7 & 77.0 & 33.3 & 40.4 & 0.0 & 0.2 \\
  \quad Embedding Cosine
      & 100.0 & 95.3 & 16.7 & 11.2 & 0.0 & 3.4 \\
  \multicolumn{7}{l}{\quad\textit{Logit-based}} \\
  \quad Raw likelihood
      & 28.6 & 37.6 & 0.0 & 2.4 & 0.0 & 0.0 \\
  \quad ZLIB-normalized
      & 100.0 & 100.0 & 0.0 & 0.8 & 0.0 & 0.0 \\
  \quad Min-$k$
      & 28.6 & 45.0 & 0.0 & 4.1 & 0.0 & 0.0 \\
\midrule
  \multicolumn{7}{l}{\textbf{\em Reference-Based Methods}} \\
  \multicolumn{7}{l}{\quad\textit{String-based}} \\
  \quad Multi n-gram Jaccard
      & 87.5 & 61.0 & 66.7 & 54.6 & 16.7 & 31.6 \\
  \quad Symmetric Coverage
      & 75.0 & 58.4 & 83.3 & 53.5 & 16.7 & 37.0 \\
  \quad Embedding Cosine
      & 50.0 & 43.3 & 66.7 & 59.2 & 0.0 & 22.9 \\
  \quad \textbf{Raw likelihood (Ours)}
      & \textbf{100.0} & \textbf{100.0} & \textbf{100.0} & \textbf{100.0} & \textbf{100.0} & \textbf{99.9} \\
\bottomrule
\end{tabular}
\caption{
    \textbf{Teacher identification accuracy (\%) in controlled experiments}.
    Accuracy is broken down by true teacher model, including both aggregated and per-sample metrics.
    Our method, which is the reference-based variant of likelihood, achieves near-perfect accuracy.
}
\label{tab:teacher_identification_controlled}
\end{table}

\begin{table}[!t]
\centering
\footnotesize
\setlength{\tabcolsep}{6pt}
\begin{tabular}{l c c c}
\toprule
True teacher & $n$ & $p$ range vs chance & Significant \\
\midrule
GPT-OSS-120B            & 6 & $3.9{\times}10^{-121}$ -- $2.3{\times}10^{-118}$ & $6/6$ \\
Qwen-3-8B               & 6 & $3.9{\times}10^{-121}$ -- $1.4{\times}10^{-113}$ & $6/6$ \\
Llama-3.3-70B-Instruct  & 7 & $3.9{\times}10^{-121}$ -- $3.5{\times}10^{-71}$  & $7/7$ \\
\midrule
\textbf{Overall}        & 19 & ---                                              & $19/19$ \\
\bottomrule
\end{tabular}
\caption{\textbf{Per-probe top-1 significance in controlled experiments.}
For each retained student, we count how often the true teacher is the top-scoring candidate
across the cross-dataset probes (s1$\to$OMI and OMI$\to$s1) and test this win rate against chance
with a one-sided exact binomial test. We report the range of $p$-values across students for each
teacher. $n$ varies because we exclude distilled students that underperform their base model. All
19 retained students are significant.}
\label{tab:top1_controlled}
\end{table}

\tightparagraph{Detection method details.}

We use the pre-distilled base model corresponding to the student $S$ as a reference model ($R$), e.g., if $S$ is obtained by distilling a base model, we use that original base model as $R$. We perform ablations on this choice. We also truncate all teacher outputs to 2,048 tokens to prevent output-length variation from trivializing detection.

As proxy prompt datasets $\hat{\mathcal{X}}$: we assume no access to the original distillation inputs $\mathcal{X}$ and use disjoint inputs, e.g., s1 for models distilled on OMI prompts, and vice versa. 
For Distilled R1 models, since the true distillation prompts are truly unknown, we use both s1 and OMI as proxy prompts ($\hat{\mathcal{X}}$).

We compare against length-matching, string-based (multi $n$-gram Jaccard, symmetric coverage, embedding cosine), and membership-inference baselines (raw likelihood, ZLIB-normalized, Min-$k$, and reverse variants). \S\ref{app:exp_details} provides the complete controlled-experiment configurations, distillation hyperparameters, and baseline definitions. We report both sorted per-sample (Per-sample) and aggregated (Agg.) accuracy, as defined in \S\ref{subsec:teacher-identification-method}.

\subsection{Results from Controlled Experiments}\label{subsec:teacher_identification_controlled}

Table~\ref{tab:teacher_identification_controlled} reports results across the 19 retained distilled students. Our reference-based method achieves near-perfect accuracy in every setting, while baselines are inconsistent---performing well only when Llama 3.3 70B Instruct is the teacher and failing on Qwen-3-8B and GPT-OSS-120B.

\begin{wrapfigure}{r}{0.4\linewidth}
    \vspace{-1em}
    \centering
    \includegraphics[width=\linewidth]{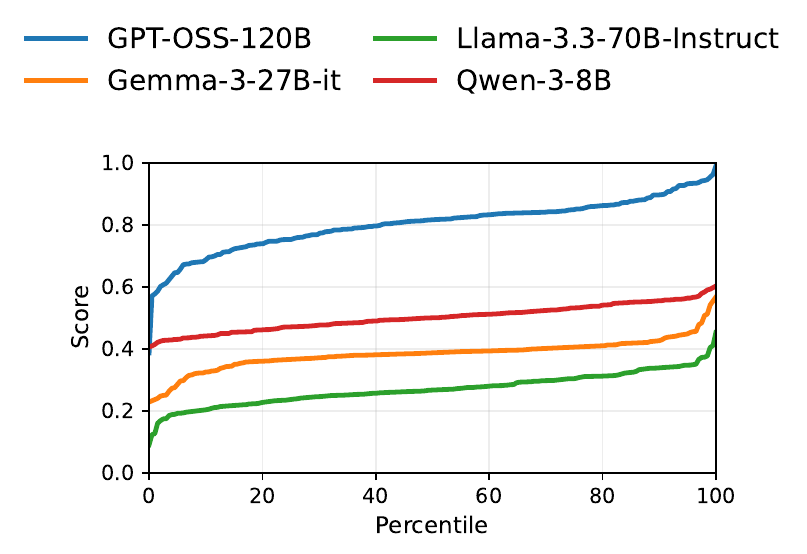}
    \caption{
    \textbf{Reference-based MIA CDFs.} Qwen-2.5-3B distilled from GPT-OSS-120B on s1, probed with OMI.}
    \vspace{-1em}
    \label{fig:ref_mia_example}
\end{wrapfigure}

Figure~\ref{fig:ref_mia_example} shows a representative controlled example: a Qwen-2.5-3B-Base student distilled from GPT-OSS-120B on s1 prompts. The true teacher is clearly separated under the non-training data probes (OMI), and we observe the same qualitative pattern across all controlled student--teacher pairs.

Statistical significance testing in Table~\ref{tab:top1_controlled} further confirms that the true teacher's per-probe advantage is not attributable to sampling noise. Across all 19 students, the true teacher significantly exceeds the $1/K = 0.25$ chance baseline under a one-sided exact binomial test ($p=3.5\times10^{-71}$ to $3.9\times10^{-121}$). Because the test evaluates a pre-specified teacher rather than selecting the best of $K$ candidates, no multiple-comparison correction is required. As an exact test, it also avoids distributional assumptions and does not require normality checks or Wilcoxon-style alternatives.

\begin{figure*}[!t]
    \centering
    \includegraphics[width=0.90\textwidth]{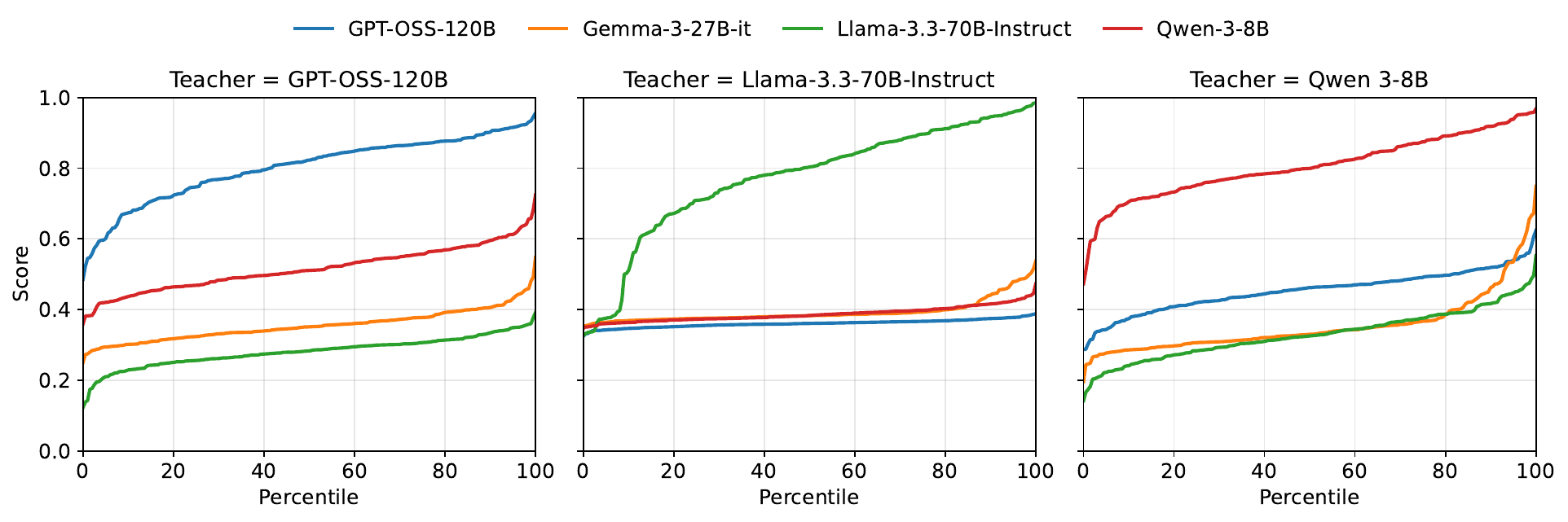}
    \caption{
    Reference-based MIA after only 32 training examples (checkpoint 2). A Qwen-2.5-1.5B student distilled from each teacher (GPT-OSS-120B, Llama-3.3-70B-Instruct, Qwen-3-8B) on OMI prompts, probed with s1 and referenced against the base Qwen-2.5-1.5B. In every panel the true teacher is the top-ranked candidate, showing that the teacher signal is already identifiable after only a few dozen examples.
    }
    \label{fig:sample_ablation}
\end{figure*}

We additionally ablate how many teacher-generated examples are needed for the reference-based signal to emerge. Starting from the same controlled setup above, we vary the number of training examples used to distill the student while keeping the probing and candidate teacher pool fixed. Checkpointing a Qwen-2.5-1.5B student every 16 training examples and re-running reference-based MIA at each stage---using the base Qwen-2.5-1.5B as the reference model---we find the signal emerges almost immediately: after just 32 examples, the true teacher is correctly identified for all three teachers (GPT-OSS-120B, Qwen-3-8B, and Nvidia-Llama-3.3-70B-Instruct), as shown in Figure~\ref{fig:sample_ablation}. The preceding checkpoint (16 examples) yields an identically zero reference-normalized loss, since under our cosine schedule with 5\% warmup the first step is taken at a near-zero learning rate, leaving the student unchanged from the base reference. Identification then remains stable through the end of training, indicating that only a few dozen teacher-generated examples suffice for the true teacher to dominate.

\tightparagraph{Effect of highly customized instructions in distillation detection}\label{para:custom-instructions}
\citet{toshniwal2024openmathinstruct2} introduce a specialized prompt format (OMI COT) that they show substantially alters Llama Instruct response style, making this a challenging setting where prompt-induced stylistic artifacts may obscure the teacher signal. Following \citet{toshniwal2024openmathinstruct2}, we prompt Llama 3.3 70B Instruct with the OMI COT template, distill students on this data, and test whether reference-based MIA recovers the teacher using s1 as probing data (since OMI was used for training). Table~\ref{tab:teacher_identification_controlled_customized_instructions} shows that detection accuracy degrades substantially under this distribution shift.

\begin{table}[t]
\centering
\footnotesize
\setlength{\abovecaptionskip}{3pt}
\setlength{\belowcaptionskip}{-4pt}
\begin{tabular}{l rr rr}
\toprule
    \multirow{2}{*}{Method}
        & \multicolumn{2}{c}{Default instructions}
        & \multicolumn{2}{c}{Highly customized instructions} \\
    \cmidrule(lr){2-3} \cmidrule(lr){4-5}
    & Agg. & Per-sample
    & Agg. & Per-sample
    \\
\midrule
    Random & 25.0 & 25.0 & 25.0 & 25.0 \\
\midrule
    \textbf{Ours} (Reference-based Raw likelihood)      &  100.0 & 100.0 & 50.0 & 65.9 \\
    ~~~~+ In-context exemplars from $S$  &  N/A   & N/A   & 75.0 & 67.1 \\
\bottomrule
\end{tabular}
\caption{Comparison of reference-based teacher identification under highly customized instructions during distillation (unknown to the detector), with and without in-context exemplars.}
\label{tab:teacher_identification_controlled_customized_instructions}
\end{table}

\begin{figure*}[!t]
    \centering
    \includegraphics[width=0.90\textwidth]{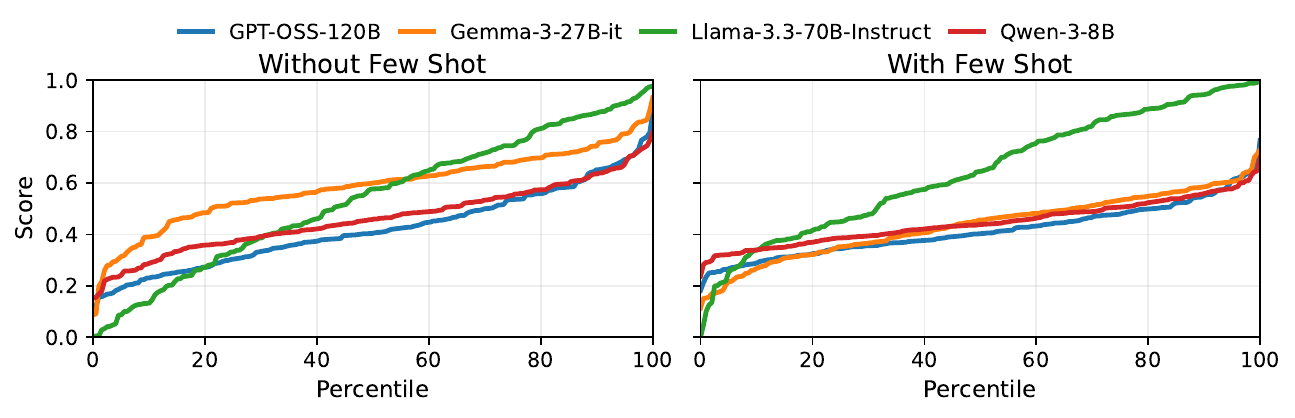}
    \caption{
    Results on a model distilled from Llama 3.3 70B Instruct, using highly customized instructions (unknown to the detector). While the default method struggles to identify the true teacher (left), few-shot prompting improves teacher identification (right).
    }
    \label{fig:omi_fewshot}
\end{figure*}

To address this, we introduce a \textbf{simple few-shot prompting strategy}: for each candidate teacher, we prepend a small number of student-generated (prompt, response) examples under the default template to better align teacher outputs with the student's style. As shown in the last row of Table~\ref{tab:teacher_identification_controlled_customized_instructions}, this improves accuracy, recovering the true teacher for three of four student models (the exception being Gemma-3-4B-pt). Figure~\ref{fig:omi_fewshot} illustrates the improved separation between the true teacher and alternatives, with additional CDFs shown in \S\ref{app:appendix-OMI-2-COT-plots}.

\S\ref{app:exp_results} reports additional ablations on reference-model choice and proxy-prompt mismatch. Performance is generally robust to reference choice, although reliable identification benefits from a reasonably similar reference; for instruction-tuned students, same-family, earlier-generation instruction-tuned references are more effective than non-instruction-tuned ones. Performance also remains stable when proxy prompts differ substantially from the original distillation inputs, including in cross-domain settings such as math versus code.

\begin{figure}[t]
\centering

\begin{minipage}[t]{0.48\textwidth}
\vspace{0pt}
\centering
\footnotesize
\setlength{\tabcolsep}{3pt}

\begin{tabular}{@{}l rr@{}}
\toprule
\multirow{2}{*}{Method}
& \multicolumn{2}{c}{Avg. over 7 models} \\
\cmidrule(lr){2-3}
& Agg. & Per-samp. \\
\midrule
Random
& -- & 10.0 \\
\midrule
\multicolumn{3}{@{}l}{\textit{Non Reference-Based Methods}} \\
\quad Multi $n$-gram Jaccard
& 33.3 & 49.5 \\
\quad Symmetric Coverage
& 16.7 & 40.3 \\
\midrule
\multicolumn{3}{@{}l}{\textit{Reference-Based Methods}} \\
\quad Multi $n$-gram Jaccard
& 100.0 & 56.8 \\
\quad Symmetric Coverage
& 83.3 & 41.1 \\
\quad Raw likelihood (R1 answers only)
& 0.0 & 0.5 \\
\quad \textbf{Raw likelihood (Ours)}
& \textbf{100.0} & \textbf{97.2} \\
\bottomrule
\end{tabular}

\vspace{.3em}
\captionsetup{type=table}
\captionof{table}{
\textbf{Teacher identification accuracy (\%) for real-world distilled
models (OMI probe set).}
Averaged over seven models: six R1-Distill models plus s1.1-32B.
}
\label{tab:teacher_identification_real_world_avg}
\end{minipage}
\hfill
\begin{minipage}[t]{0.50\textwidth}
\vspace{0pt}
\centering
\vspace{0.25em}
\includegraphics[
    width=\linewidth
]{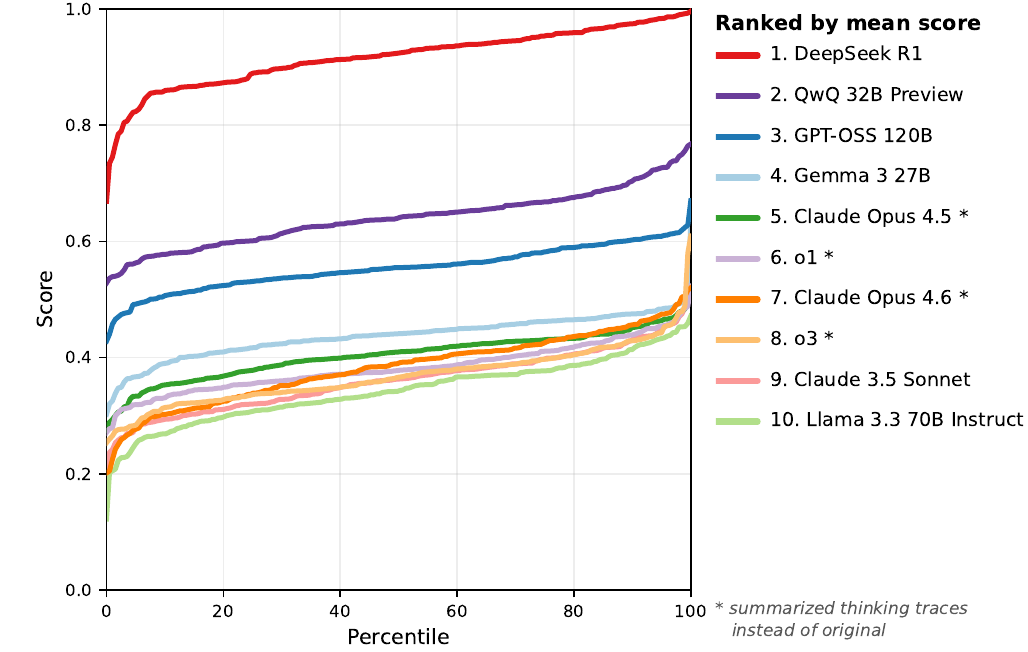}
\vspace{-1.07em}
\caption{
Results for \texttt{s1.1-32B}, a real-world model trained by
\cite{muennighoff2025s1}. The true teacher model, R1, is ranked at the
top by a large margin.
}
\label{fig:s11_xcoder_combined}
\end{minipage}
\end{figure}

\subsection{Results on Real-World Distilled Models
}\label{subsec:teacher-identification-models-in-the-wild-results}
\paragraph{Results on DeepSeek-R1 distill models and s1.1-32B}
As reported in Table~\ref{tab:teacher_identification_real_world_avg}, we find our method correctly identifies the true teacher among ten candidates for seven out of the seven evaluated models, both over OMI and s1 probing sets, showing that our method works when distillation data and training procedure are largely unknown.

The same per-probe test confirms this at the level of
individual probes (Table~\ref{tab:top1_wild}): for all seven targets, on both prompt
sets, DeepSeek-R1's per-probe win rate is statistically significantly above the $1/K$
chance baseline. As in the controlled setting, the teacher is named in advance, so the
test carries no best-of-$K$ selection penalty, certifying that this per-probe lead is
not sampling noise.
 
\begin{table}[t]
\centering \vspace{-.5em}
\footnotesize
\setlength{\tabcolsep}{8pt}
\begin{tabular}{l c}
\toprule
True teacher & $p$ range vs chance \\
\midrule
DeepSeek-R1 & $1.0\times10^{-200}$ -- $3.4\times10^{-138}$ \\
\bottomrule
\end{tabular}
\caption{\textbf{Per-probe top-1 significance on real-world distilled models.}
The true teacher is DeepSeek-R1 for all seven targets. For each, we count how often R1 ranks first
across both prompt sets against the full candidate pool and test against chance with a one-sided
exact binomial test. We report the range of $p$-values across targets and prompt sets. All seven
targets are significant on both prompt sets.}
\label{tab:top1_wild}
\end{table}

The CDF graph for \texttt{s1.1-32B} is provided in Figure~\ref{fig:s11_xcoder_combined} for reference, showing that the true teacher (R1) is clearly separated from other candidates. 

Additional CDFs for the remaining real-world models, including the \texttt{X-Coder-SFT-Qwen3-8B} results in Figure~\ref{fig:xcoder-ref}, are provided in \S\ref{app:appendix-models-in-the-wild}.

\tightparagraph{Role of Reasoning Traces.}
One important takeaway from these experiments is that access to teacher reasoning traces is critical for reliable teacher identification. As shown in the bottom section of Table~\ref{tab:teacher_identification_real_world_avg}, when R1 reasoning traces are hidden and only final answers are available, teacher identification accuracy collapses to zero. This suggests an important limitation for settings where full teacher reasoning traces are unavailable, such as closed reasoning models that expose only final answers or compressed summaries of their reasoning.

\FloatBarrier
\section{Experiment: Distillation Detection}\label{sec:distillation-detection}

This section evaluates how the margin-threshold detector from
\S\ref{subsec:distillation-detection-method} generalizes beyond the models used
to calibrate it. Selecting $\tau$ to maximize controlled accuracy yields
$\tau=0.067$, which attains $85.5\%$ $(65/76)$ on the controlled set but only
$53.6\%$ $(15/28)$ when applied unchanged to the real-world distilled cells---a
single fixed threshold does not transfer cleanly between settings. We therefore
isolate two forms of generalization: transfer to an unseen \emph{student} whose
teacher was seen during calibration and transfer to an unseen \emph{teacher}.
For each, we report the threshold's accuracy and the significance of every
detection.

\subsection{Experimental Setup}

\paragraph{Data Construction.}
Each evaluation cell is $(S,\{T_1,\ldots,T_K\},y)$, where $y=1$ if the true
teacher is present in the candidate pool and $y=0$ otherwise. We use two sources:

\begin{itemize}[leftmargin=14pt, topsep=1pt, itemsep=0pt]

    \item \textbf{Controlled models.}
    The 19 distilled students from
    \S\ref{subsec:teacher-identification-setup}, each evaluated with our
    Reference MIA method under OMI and s1 prompts with the true teacher included
    ($y{=}1$, $K{=}4$) or excluded ($y{=}0$, $K{=}3$), yielding
    $19\times2\times2=76$ evaluation cells.

    \item \textbf{Real-world distilled models.}
    Six DeepSeek-R1 distilled models and s1.1-32B, each evaluated under both
    prompt datasets with the true teacher included ($y{=}1$, $10$ candidates)
    or excluded ($y{=}0$, $9$ candidates), yielding
    $7\times2\times2=28$ held-out cells.

\end{itemize}
\tightparagraph{Cross-Validation.}
In every cross-validation fold, we refit the detection threshold $\tau$ using only the training cells, selecting the value that maximizes F1 without access to any held-out margins.
For the controlled models, we use two splits. In
\emph{leave-one-student-out}, we hold out one student and fit $\tau$ on
the rest, leaving all teachers represented (four folds).
In \emph{leave-one-teacher-out}, we hold out every distillation of one teacher,
leaving that teacher absent during calibration (three folds). For the
real-world distilled models, all seven share DeepSeek-R1 as their teacher, so we only perform leave-one-student-out, fitting $\tau$ on six R1-distilled students and evaluating the seventh. For each held-out cell, we report margin-prediction accuracy ($\delta\ge\tau$) and apply the detection test from \S\ref{subsec:significance} with the fold's $\tau$ in place of $\tau^*$.

\subsection{Results}
\label{subsec:distillation-detection-results}

\paragraph{Generalization to unseen students versus unseen teachers.}
Table~\ref{tab:threshold-generalization-controlled} breaks the controlled
cross-validation down by true teacher. The two reasoning teachers
(GPT-OSS-120B and Qwen-3-8B) are detected for every one of their students
($6/6$ each) at $p^*<10^{-20}$ under both splits, whereas the non-reasoning
Llama-3.3-70B-Instruct is detected for only $1/7$ students regardless of the
split---its margin is too faint to clear $\tau$. The splits diverge in
specificity, not detection: leaving a student out produces no false detections,
whereas leaving a teacher out produces false detections on the reasoning
teachers' teacher-removed cells ($4/6$ and $1/6$), because the unseen teacher's
regime mis-sets $\tau$. Overall, accuracy falls from $82.9\%$ $(63/76)$---close
to the $85.5\%$ in-sample fit---under leave-one-student-out to $60.5\%$
$(46/76)$ under leave-one-teacher-out.

\begin{table}[!t]
\centering
\footnotesize
\setlength{\tabcolsep}{4.5pt}
\begin{tabular}{l c c c c c c}
\toprule
True teacher
    & Detected
    & Detection $p^*$
    & Acc.\ (S-out)
    & Acc.\ (T-out)
    & FP (S-out)
    & FP (T-out) \\
\midrule
GPT-OSS-120B
    & $6/6$
    & ${<}10^{-26}$
    & $24/24$
    & $15/24$
    & $0/6$
    & $4/6$ \\
Qwen-3-8B
    & $6/6$
    & ${<}10^{-20}$
    & $22/24$
    & $15/24$
    & $0/6$
    & $1/6$ \\
Llama-3.3-70B-Instruct
    & $1/7$
    & $6{\times}10^{-34}$--$1.0$
    & $17/28$
    & $16/28$
    & $0/7$
    & $0/7$ \\
\midrule
\textbf{Overall}
    & $13/19$
    & ---
    & $63/76$
    & $46/76$
    & $0/19$
    & $5/19$ \\
\bottomrule
\end{tabular}
\caption{\textbf{Controlled cross-validation by true teacher.}
\emph{Detected} and \emph{Detection $p^*$} are the detection test on the true-teacher-present cells,
using the leave-one-student-out fold $\tau$; \emph{FP} is a substitute
falsely called when the true teacher is removed, under the S-out $\tau$
(FP S-out) or T-out $\tau$ (FP T-out). \emph{Acc.}\ is the margin prediction
($\delta\ge\tau$) per cell with the corresponding fold $\tau$.
S-out $=$ leave-one-student-out, T-out $=$ leave-one-teacher-out.}
\label{tab:threshold-generalization-controlled}
\end{table}

\tightparagraph{Generalization across real-world students.}
The real-world experiment reproduces the student-side pattern
(Table~\ref{tab:threshold-generalization-wild}). With $\tau$ refit on the other
six R1-distilled models in each fold, leave-one-student-out attains $92.9\%$
$(26/28)$ accuracy. The detection test calls six of seven held-out students
distilled from DeepSeek-R1 (both prompts, $p^*$ between
$1.2{\times}10^{-26}$ and $8.0{\times}10^{-3}$); when DeepSeek-R1 is removed,
no alternative ever clears the threshold ($p^*\ge0.078$), giving zero false
detections. The single missed positive is s1.1-32B: DeepSeek-R1 ranks first on
both prompts, but its OMI margin ($0.105$) falls below the fold-fit threshold
($\tau=0.176$), so OMI is non-significant even though s1 is overwhelming
($p^*=8.5{\times}10^{-34}$), and the both-prompts rule withholds the call.

\begin{table}[!t]
\centering
\footnotesize \vspace{-1em}
\setlength{\tabcolsep}{8pt}
\begin{tabular}{l c c c}
\toprule
Held-out condition & Accuracy & Detected & Detection $p^*$ \\
\midrule
DeepSeek-R1 included
    & $13/14$
    & $6/7$
    & $1.2{\times}10^{-26}$--$8.0{\times}10^{-3}$ \\
DeepSeek-R1 removed
    & $13/14$
    & $0/7$
    & $\ge 0.078$ (n.s.) \\
\midrule
\textbf{Overall}
    & $\mathbf{26/28}$ $(92.9\%)$
    & ---
    & --- \\
\bottomrule
\end{tabular}
\caption{\textbf{Leave-one-student-out (real-world).}
Margin accuracy and significant detections (both prompts, same teacher); with
DeepSeek-R1 removed, no alternative is significant.}
\label{tab:threshold-generalization-wild}
\end{table}

\tightparagraph{Why teacher transfer is harder.}
The scale of the reference-normalized signal depends far more on the teacher
than on the student. A threshold calibrated on the strong signals of the reasoning teachers is thus consistent
across their students, however a non-reasoning teacher may leave a margin too faint to
clear it as shown with Llama 3.3 70B Instruct in Table~\ref{tab:threshold-generalization-wild}, while a threshold pulled lower by that teacher admits false
detections on the reasoning teachers' teacher-removed cells---the specificity
loss seen under leave-one-teacher-out. Our results rest on only three controlled teachers,
so the mechanism is suggestive rather than conclusive; nonetheless, the
cross-validation supports a clear conclusion: detection generalizes to unseen
students when their teacher is represented during calibration, while threshold
calibration is less reliable across unseen teachers.

\FloatBarrier
\section{OpenAI o1 Detection Methods}\label{sec:O1detection}
\paragraph{Method.}
We serialize each \texttt{o1} output in two ways: Unicode, preserving raw UTF-8
characters, and ASCII, escaping each non-ASCII codepoint. We then run our reference-MIA
method with all other hyperparameters fixed and define
$\Delta_{\text{ASCII}} = \overline{\mathcal{L}}_{\text{ASCII}} - \overline{\mathcal{L}}_{\text{Unicode}}$,
where positive values indicate that ASCII normalization raises reference-normalized loss.
We evaluate four \texttt{o1}-distilled students trained on s1 using 100 held-out OMI
probes, and compare against real U.S.-organization control models not expected to be
distilled from \texttt{o1}/\texttt{o3}. To assess significance we reuse the
statistical test (\S\ref{subsec:significance}) with the null placed at zero;
additional details, controls, and wild-model results are in \S\ref{app:UnicodeResults}.

\begin{figure}[H]
\centering \vspace{-1em}
\includegraphics[width=0.64\linewidth]{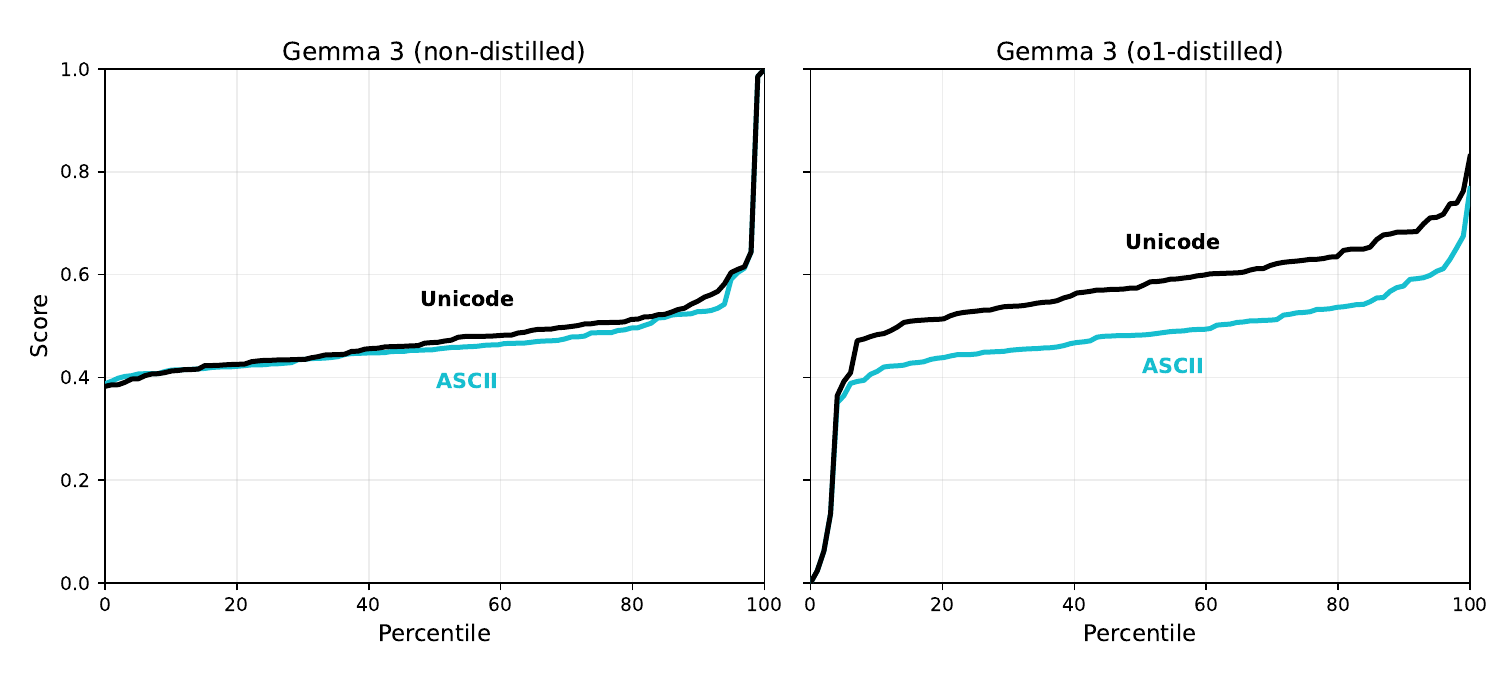}
\vspace{-1em}
\caption{
\textbf{Per-probe CDFs.}
CDFs for a within-family control and an \texttt{o1}-distilled Gemma student. The
non-distilled \texttt{gemma-3-27b-pt} control has nearly overlapping ASCII and Unicode
curves, whereas the \texttt{o1}-distilled \texttt{gemma-3-4b-pt} student separates
consistently.
}\vspace{-1em}
\label{fig:o1_cdf}
\end{figure}

\begin{table}[H]
\centering
\footnotesize
\setlength{\tabcolsep}{5pt}
\begin{tabular}{l r c r c}
\toprule
Target & $\Delta_{\text{ASCII}}$ & $95\%$ CI & $p$ & Sig. \\
\midrule
\rowcolor{red!8} \texttt{gemma-3-4b-pt} (SFT)        & $+0.0732$ & $[+0.064,+0.083]$ & ${<}10^{-15}$ & \cmark \\
\rowcolor{red!8} \texttt{Qwen2.5-1.5B} (SFT)         & $+0.0594$ & $[+0.050,+0.069]$ & ${<}10^{-15}$ & \cmark \\
\rowcolor{red!8} \texttt{Qwen2.5-3B} (SFT)           & $+0.0573$ & $[+0.048,+0.067]$ & ${<}10^{-15}$ & \cmark \\
\rowcolor{red!8} \texttt{Llama-3.2-3B-Instruct} (SFT)& $+0.0647$ & $[+0.055,+0.075]$ & ${<}10^{-15}$ & \cmark \\
\midrule
\rowcolor{blue!8} \texttt{gemma-3-12b-it}     & $+0.0373$ & $[+0.028,+0.047]$ & $5.0{\times}10^{-11}$ & \cmark \\
\rowcolor{blue!8} \texttt{gemma-3-27b-it}     & $+0.0072$ & $[-0.001,+0.015]$ & $2.7{\times}10^{-2}$  & \cmark \\
\rowcolor{blue!8} \texttt{gemma-3-27b-pt}     & $+0.0070$ & $[+0.003,+0.011]$ & $5.1{\times}10^{-4}$  & \cmark \\
\rowcolor{blue!8} \texttt{gemma-2-9b}         & $+0.0016$ & $[-0.005,+0.008]$ & $0.28$ & \xmark \\
\rowcolor{blue!8} \texttt{Llama-3.3-70B}      & $+0.0007$ & $[-0.005,+0.006]$ & $0.66$ & \xmark \\
\rowcolor{blue!8} \texttt{Llama-3.1-8B}       & $-0.0024$ & $[-0.007,+0.002]$ & $0.63$ & \xmark \\
\rowcolor{blue!8} \texttt{Llama-3.1-70B}      & $-0.0050$ & $[-0.008,-0.002]$ & $1.00$ & \xmark \\
\rowcolor{blue!8} \texttt{Llama-3.1-8B-Inst.} & $-0.0078$ & $[-0.017,+0.001]$ & $0.93$ & \xmark \\
\rowcolor{blue!8} \texttt{gemma-2-9b-it}      & $-0.0733$ & $[-0.105,-0.043]$ & $1.00$ & \xmark \\
\bottomrule
\end{tabular}

\caption{\textbf{ASCII vs.\ Unicode serialization with significance.}
Targets ranked by $\Delta_{\text{ASCII}}$; positive values indicate higher loss under ASCII normalization. We report $\Delta_{\text{ASCII}}$ with $95\%$ bootstrap CIs; \cmark\ denotes statistical significance. All four \texttt{o1}-distilled students are significant, as are three Gemma-3 controls---but with much smaller gaps. Red:
\texttt{o1}-distilled positives; blue: controls.}
\label{tab:o1_significance}
\vspace{-1em}
\end{table}

\tightparagraph{Results.}
Figure~\ref{fig:o1_cdf} and Table~\ref{tab:o1_significance} show that
\texttt{o1}-distilled students have consistently positive ASCII gaps, while controls
mostly remain near zero. The CDFs show the same distributional effect: the non-distilled
\texttt{gemma-3-27b-pt} control has nearly overlapping ASCII and Unicode curves, whereas
for the \texttt{o1}-distilled \texttt{gemma-3-4b-pt} student the Unicode curve lies
consistently above the ASCII curve. This ordering---Unicode assigned
higher membership likelihood than ASCII for the distilled model but not the control---may
itself serve as an additional visual indicator of distillation, mirroring the positive
$\Delta_{\text{ASCII}}$ gap. To
confirm these gaps are not sampling noise, we test each per-probe difference against
zero.\footnote{Because Unicode is fixed as the reference serialization in advance, this is a
single pre-specified directional comparison ($K=2$), so we test against zero rather than
the calibrated margin $\tau^\star$ and need no candidate-selection (Bonferroni)
correction.}
All four \texttt{o1}-distilled students have significantly positive gaps
($\Delta_{\text{ASCII}}\in[0.0573,0.0732]$; $p<10^{-15}$). Only the three Gemma-3
controls also reach significance, but with much smaller $\Delta_{\text{ASCII}}$ gaps
(\texttt{gemma-3-12b-it} $+0.0373$; \texttt{gemma-3-27b-it} $+0.0072$;
\texttt{gemma-3-27b-pt} $+0.0070$); the two Gemma-2 controls do not. With only $100$
probes, a $p$-value can resolve even a tiny positive $\Delta_{\text{ASCII}}$ from zero,
so significance alone may be a weak criterion here; what matters is the size of
$\Delta_{\text{ASCII}}$ itself---the ASCII-minus-Unicode difference in
reference-normalized loss. By that measure the distilled and control gaps are clearly
separated: the smallest distilled gap ($+0.0573$) exceeds the largest control gap
($+0.0373$) by over $50\%$ and the \texttt{27b} gaps by roughly $8\times$, and all four
distilled models' confidence intervals lie entirely above every control model's interval.
Thus, Unicode features inherited from \texttt{o1} can serve as a complementary membership signal under controlled distillation. However, because this diagnostic uses only $100$ OMI probes from a single prompt set, we view it as a cue rather than standalone evidence.

\section{Open Questions}\label{sec:openquestions}
We explore the applicability of reference-based MIA to open-ended settings where it is unknown whether a model was distilled from a third-party model. Specifically, we use reference-based MIA to probe three open-ended attribution questions:
\begin{enumerate}[leftmargin=14pt, topsep=1pt, itemsep=0pt]
    \item Is \textbf{QwQ-32B} possibly distilled from Llama, o1/o3, Claude, or DeepSeek-R1?
    \item Is \textbf{GPT-OSS} possibly distilled from Llama, QwQ, o1/o3, Claude, or DeepSeek-R1?
    \item Is \textbf{DeepSeek-R1} possibly distilled from Llama, QwQ, o1/o3, or Claude?
\end{enumerate}

\tightparagraph{Disclaimer.}
Our validation relies on several simplifying assumptions---single-teacher, SFT-style final-stage distillation---and limited experimental evaluations. Real-world models may instead be built on multiple teachers across multiple training stages. Moreover, the candidate teacher models we consider are themselves likely deeply connected, often building on one another through complex dependency chains~\citep{adhikesaven2026modelsmodelsbuilton}.
Therefore, the results in this section should be interpreted as a preliminary exploration rather than evidence sufficient to support definitive conclusions.

\tightparagraph{Candidate pools and references.}
For each target, we use the same candidate-teacher pool as in our models-in-the-wild experiments (\S\ref{subsec:teacher-identification-models-in-the-wild-results}), excluding near-variants of the target itself to avoid trivial same-family matches. Concretely, we exclude \texttt{GPT-OSS-120B} when evaluating \texttt{GPT-OSS-20B}, and exclude \texttt{QwQ-32B-Preview} when evaluating \texttt{QwQ-32B}. We use s1 and OpenMathInstruct (OMI) as probing datasets throughout.
For each experiment, we include one representative figure; additional plots and detailed statistical analyses are provided in \S\ref{app:WildModels}.

\begin{figure*}[h]
    \centering
    \vspace{-.5em}
    \includegraphics[width=1.0\textwidth]{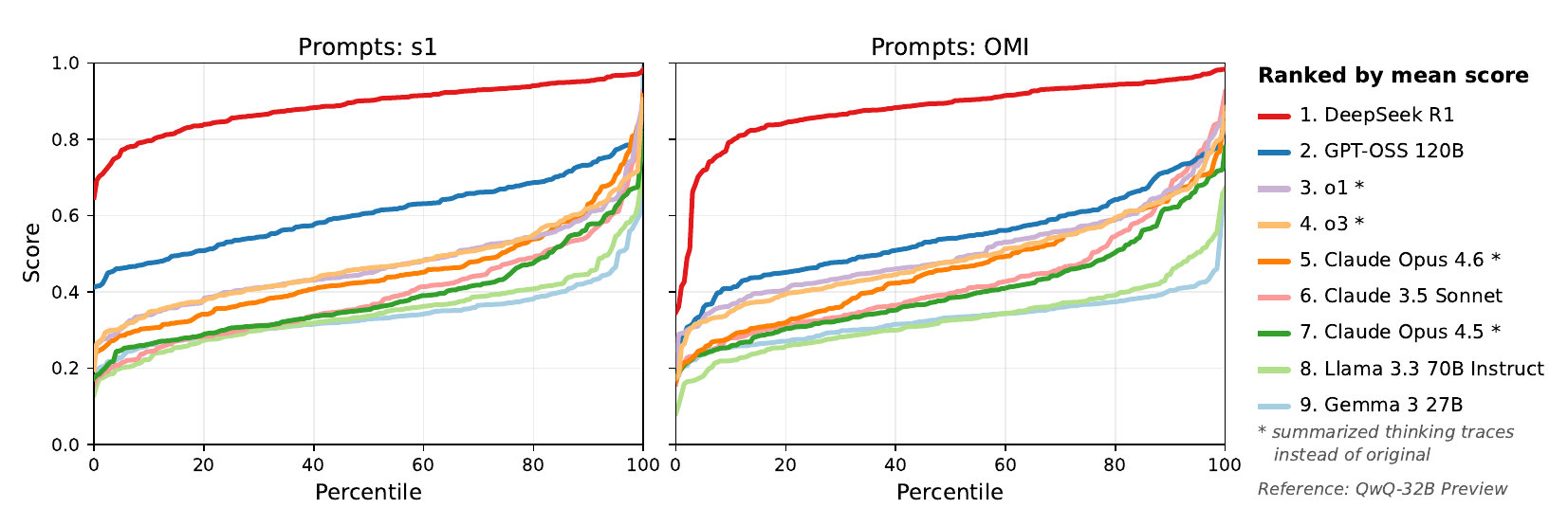}
    \vspace{-2em}
    \caption{
    Our reference-based method applied to {\fontfamily{lmtt}\fontseries{b}\selectfont QwQ-32B}, using \texttt{QwQ-32B-Preview} as a reference. Results may suggest \texttt{QwQ-32B} was distilled from R1 after the release of \texttt{QwQ-32B-Preview}.
    }\vspace{-.1em}
    \label{fig:qwq_open}
\end{figure*}
\tightparagraph{QwQ-32B.}
We evaluate \texttt{QwQ-32B} %
using \texttt{QwQ-32B-Preview} as the reference model. As shown in Figure~\ref{fig:qwq_open}, \texttt{DeepSeek-R1} ranks highest across both probing sets, with a pronounced gap from the next candidate. Combined with the release timeline---\texttt{QwQ-32B-Preview} in November 2024, \texttt{DeepSeek-R1} in January 2025, and \texttt{QwQ-32B} in March 2025---this raises the possibility that \texttt{DeepSeek-R1} outputs were used to train \texttt{QwQ-32B} following the release of \texttt{QwQ-32B-Preview}. %

\begin{figure}[h]
    \centering
    \vspace{-.5em}
    \includegraphics[width=1.0\linewidth]{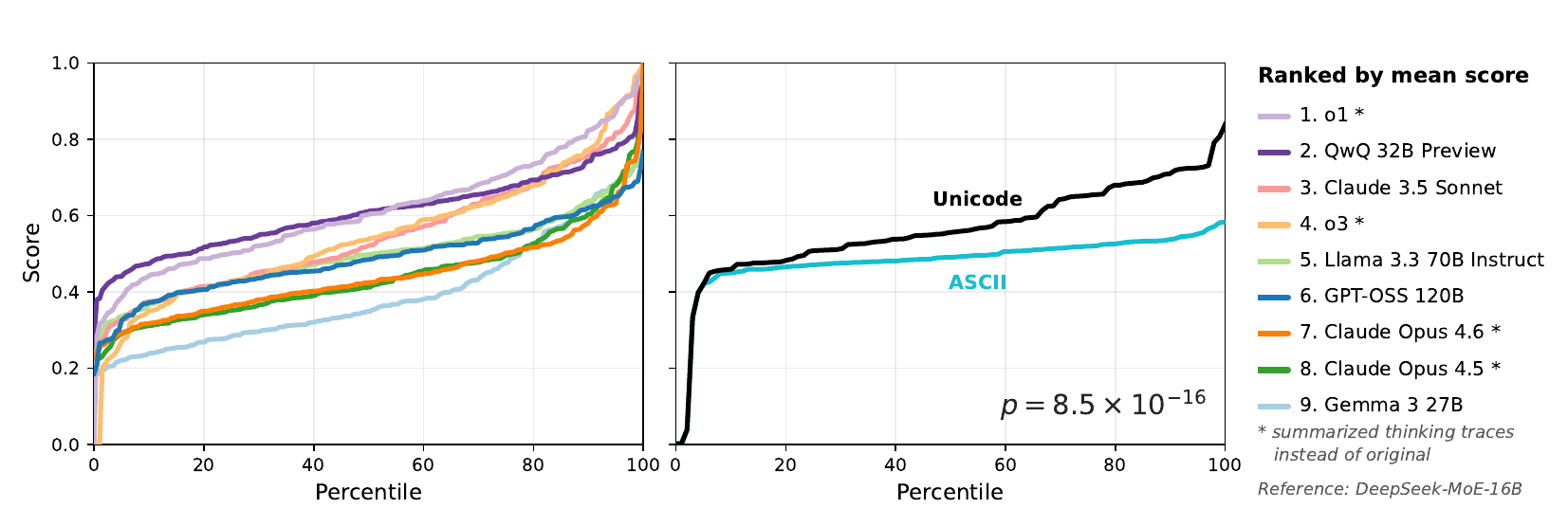}
    \vspace{-1.0em}
    \caption{Our reference-based method and ASCII-Unicode diagnostic applied to {\fontfamily{lmtt}\fontseries{b}\selectfont DeepSeek-R1} using \texttt{DeepSeek-MoE-16B} as a reference.}
    \label{fig:r1_combined_wild}
    \vspace{-.3em}
\end{figure}

\tightparagraph{DeepSeek R1.}
We evaluate \texttt{DeepSeek-R1}, using \texttt{DeepSeek-MoE-16B} as the reference model.
Results are reported in Figure~\ref{fig:r1_combined_wild}. %
Our reference-based method ranks \texttt{o1} and \texttt{QwQ-32B-Preview} near the top, raising the possibility that \texttt{DeepSeek-R1} was influenced by or distilled from \texttt{o1} or \texttt{QwQ-32B-Preview}.

In addition, the \texttt{o1} diagnostic in Figure~\ref{fig:r1_combined_wild} (right) shows a clear Unicode--ASCII gap. We test this gap one-sided against zero ($95\%$ bootstrap CI over $100$ probes; Wilcoxon, as the differences are non-normal): it is $\delta_{\text{Uni}-\text{ASCII}}=+0.937$ ($[+0.746,+1.141]$, $p=8.5\times10^{-16}$; Table~\ref{tab:o1_sig}), confirming the separation is resolvable rather than a sampling artifact. This suggests possible \texttt{o1}-style signal in \texttt{DeepSeek-R1}. %

\begin{figure}[h]
    \centering
    \vspace{-1.0em}
    \includegraphics[width=1.0\linewidth]{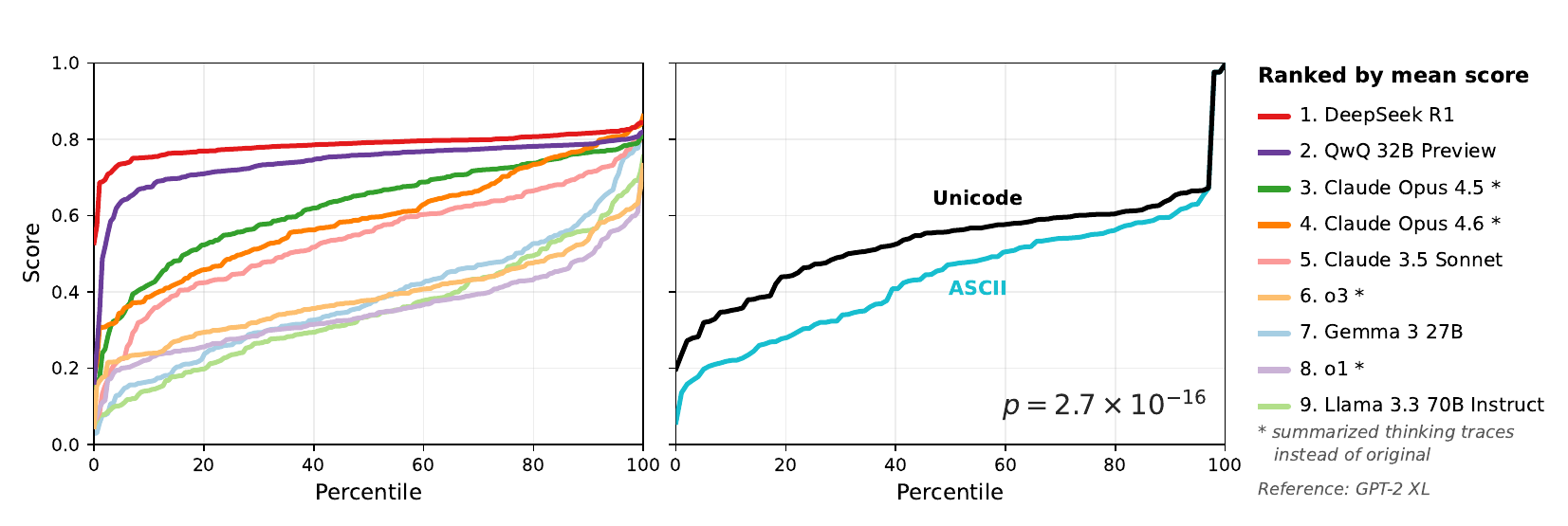}
    \vspace{-1.5em}
    \caption{Our reference-based method and ASCII-Unicode diagonistic applied to {\fontfamily{lmtt}\fontseries{b}\selectfont GPT-OSS-120B} using \texttt{GPT-2-XL} as the reference.}
    \vspace{-.3em}
    \label{fig:gptoss_combined_plot2}
\end{figure}

\tightparagraph{GPT-OSS-20B \& GPT-OSS-120B.}
We evaluate \texttt{GPT-OSS-20B \& GPT-OSS-120B}. Unlike \texttt{QwQ} and \texttt{DeepSeek-R1}, \texttt{GPT-OSS} does not have a suitable same-family checkpoint that can serve as a natural reference model. We therefore use \texttt{GPT-2 XL}, an older open-source LM released by OpenAI, while noting that it is unlikely to be a reliable reference given the substantial temporal and architectural gap. We include this evaluation primarily to illustrate how our method behaves in the absence of an appropriate reference model. Results with alternative reference models are reported in \S\ref{app:WildModels}.

As shown in Figure~\ref{fig:gptoss_combined_plot2}, \texttt{DeepSeek-R1} and \texttt{QwQ-32B-Preview} are ranked highest. However, these results are difficult to interpret, as our earlier analyses suggest that \texttt{DeepSeek-R1} and \texttt{QwQ-32B-Preview} may themselves be influenced by \texttt{o1}, which is also a plausible teacher for \texttt{GPT-OSS}.\footnote{Note that \texttt{o1} is included in our candidate teacher models but ranks low. However, this does not rule out the possibility that \texttt{GPT-OSS} was influenced by \texttt{o1}, because \texttt{o1} does not expose its original thinking traces, and this systematically underestimates the likelihood assigned to \texttt{o1}.}
An additional \texttt{o1} diagnostic (Figure~\ref{fig:gptoss_combined_plot2}, right) shows that the Unicode CDF lies consistently above the ASCII CDF, and this gap is statistically resolvable across all reference models for \texttt{GPT-OSS} (Table~\ref{tab:o1_sig}). Together, these observations raise the possibility that \texttt{GPT-OSS} is influenced by the same \texttt{o1}/\texttt{DeepSeek-R1}/\texttt{QwQ} lineage. However, because \texttt{GPT-OSS} lacks a suitable reference model, this conclusion remains highly uncertain. We leave a more systematic analysis to future work that relaxes the assumption that an appropriate reference model is available.

\FloatBarrier
\section{Conclusion}\label{sec:conclusion}

We introduce reference based distillation detection: identifying the teacher used to train a model given an earlier-generation checkpoint from the same lineage. By comparing normalized likelihood differences between a student and reference model, our approach identifies candidate teachers whose outputs are preferentially aligned with the student. Across controlled and real-world settings---including DeepSeek-R1 distills and s1.1---our method consistently identifies the true teacher and outperforms similarity-based baselines, while remaining effective under cross-domain evaluation. Overall, our work provides an initial step toward practical auditing of distillation in modern LLMs.

\tightparagraph{Limitations.}
Our evaluation targets a canonical setting---single-teacher, SFT-style, final-stage distillation---and does not address multi-teacher or multi-stage pipelines. Identification quality depends on having a suitable reference checkpoint, which is not always available, as illustrated by \texttt{GPT-OSS}. Our signal also performs well with reasoning traces of candidate teachers, so the method is most reliable for teachers that expose full traces. A further question we do not resolve concerns \emph{rewriting} pipelines: if data is generated by teacher $A$ and then paraphrased or rewritten by teacher $B$ before training, it is unclear which model our method would attribute the student to---$A$, $B$, some mixture, or neither. The detected signal could plausibly track either the originating model or the rewriter depending on how much stylistic and distributional structure survives the rewrite, and we leave a systematic study of such to future work.

\section*{Acknowledgement}
We thank David Wagner and the SM group members for valuable discussion and feedback. This work was supported in part by gifts from OpenAI, Ai2, and Apple.

\bibliographystyle{plainnat}
\bibliography{references}

@article{hinton2015distilling,
  title={Distilling the Knowledge in a Neural Network},
  author={Hinton, Geoffrey and Vinyals, Oriol and Dean, Jeff},
  journal={arXiv preprint arXiv:1503.02531},
  year={2015}
}

@article{wang2022selfinstruct,
  title={Self-Instruct: Aligning Language Models with Self-Generated Instructions},
  author={Wang, Yizhong and Kordi, Yeganeh and Mishra, Swaroop and Liu, Alisa and Smith, Noah A. and Khashabi, Daniel and Hajishirzi, Hannaneh},
  journal={arXiv preprint arXiv:2212.10560},
  year={2022}
}

@article{toshniwal2024openmathinstruct2,
  title={OpenMathInstruct-2: Accelerating AI for Math with Massive Open-Source Instruction Data},
  author={Toshniwal, Shubham and Du, Wei and Moshkov, Ivan and Kisacanin, Branislav and Ayrapetyan, Alexan and Gitman, Igor},
  journal={arXiv preprint arXiv:2410.01560},
  year={2024}
}

@article{muennighoff2025s1,
  title={s1: Simple Test-Time Scaling},
  author={Muennighoff, Niklas and Yang, Zitong and Shi, Weijia and Li, Xiang Lisa and Fei-Fei, Li and Hajishirzi, Hannaneh and Zettlemoyer, Luke and Liang, Percy and Cand{\`e}s, Emmanuel and Hashimoto, Tatsunori},
  journal={arXiv preprint arXiv:2501.19393},
  year={2025}
}

@inproceedings{shokri2017membership,
  title={Membership Inference Attacks Against Machine Learning Models},
  author={Shokri, Reza and Stronati, Marco and Song, Congzheng and Shmatikov, Vitaly},
  booktitle={IEEE Symposium on Security and Privacy},
  year={2017}
}

@inproceedings{carlini2022lira,
  title={Membership Inference Attacks From First Principles},
  author={Carlini, Nicholas and Chien, Steve and Nasr, Milad and Song, Shuang and Terzis, Andreas and Tramer, Florian},
  booktitle={IEEE Symposium on Security and Privacy},
  year={2022}
}

@inproceedings{kim2016sequence,
  title={Sequence-Level Knowledge Distillation},
  author={Kim, Yoon and Rush, Alexander M},
  booktitle={Proceedings of the 2016 Conference on Empirical Methods in Natural Language Processing},
  year={2016}
}

@article{sanh2019distilbert,
  title={DistilBERT, a distilled version of BERT: smaller, faster, cheaper and lighter},
  author={Sanh, Victor and Debut, Lysandre and Chaumond, Julien and Wolf, Thomas},
  journal={arXiv preprint arXiv:1910.01108},
  year={2019}
}

@inproceedings{jiao2019tinybert,
  title={TinyBERT: Distilling BERT for Natural Language Understanding},
  author={Jiao, Xiaoqi and Yin, Yichun and Lifeng, Shang and Jiang, Xin and Chen, Xiao and Li, Linlin and Wang, Fang and Liu, Qun},
  booktitle={Findings of the Association for Computational Linguistics: EMNLP 2020},
  year={2020}
}

@article{shi2025knowledge,
  title={Knowledge Distillation Detection for Open-weights Models},
  author={Shi, Qin and Zheng, Amber Yijia and Song, Qifan and Yeh, Raymond A},
  journal={arXiv preprint arXiv:2510.02302},
  year={2025}
}

@misc{maini2024llmdatasetinferencedid,
      title={LLM Dataset Inference: Did you train on my dataset?}, 
      author={Pratyush Maini and Hengrui Jia and Nicolas Papernot and Adam Dziedzic},
      year={2024},
      eprint={2406.06443},
      archivePrefix={arXiv},
      primaryClass={cs.LG},
      url={https://arxiv.org/abs/2406.06443}, 
}

@misc{googledistillation,
      title={Google says attackers used 100,000+ prompts to try to clone AI chatbot Gemini}, 
      author={NBC News},
      year={2026},
      url={https://www.nbcnews.com/tech/security/google-gemini-hit-100000-prompts-cloning-attempt-rcna258657}, 
}

@misc{anthropicdistillation,
      title={Detecting and preventing distillation attacks}, 
      author={Anthropic},
      year={2026},
      url={https://www.anthropic.com/news/detecting-and-preventing-distillation-attacks}, 
}

@misc{openaidistillation,
      title={OpenAI says China's DeepSeek trained its AI by distilling US models, memo shows}, 
      author={Seetharaman, Deepa and Arámburo, Fabiola},
      year={2026},
      url={https://www.reuters.com/world/china/openai-accuses-deepseek-distilling-us-models-gain-advantage-bloomberg-news-2026-02-12/}, 
}

@article{abdin2024phi,
  title={Phi-4 technical report},
  author={Abdin, Marah and Aneja, Jyoti and Behl, Harkirat and Bubeck, S{\'e}bastien and Eldan, Ronen and Gunasekar, Suriya and Harrison, Michael and Hewett, Russell J and Javaheripi, Mojan and Kauffmann, Piero and others},
  journal={arXiv preprint arXiv:2412.08905},
  year={2024}
}

@article{taori2023alpaca,
  title={Alpaca: A strong, replicable instruction-following model},
  author={Taori, Rohan and Gulrajani, Ishaan and Zhang, Tianyi and Dubois, Yann and Li, Xuechen and Guestrin, Carlos and Liang, Percy and Hashimoto, Tatsunori B},
  journal={Stanford Center for Research on Foundation Models. https://crfm. stanford. edu/2023/03/13/alpaca. html},
  volume={3},
  number={6},
  pages={7},
  year={2023}
}

@article{philipp2026prompt,
  title={From Prompt to Clone: Copyright Challenges in AI Model Distillation},
  author={Philipp, Claudia},
  journal={UC Law Science and Technology Journal},
  volume={17},
  number={1},
  pages={49},
  year={2026}
}

@misc{meta_llama32_3b_instruct,
  title        = {meta-llama/Llama-3.2-3B-Instruct},
  author       = {{Meta}},
  howpublished = {\url{https://huggingface.co/meta-llama/Llama-3.2-3B-Instruct}},
  note         = {Hugging Face model card, accessed 2026-04-05}
}

@article{gemma3_technical_report,
  title   = {Gemma 3 Technical Report},
  author  = {{Gemma Team}},
  journal = {arXiv preprint arXiv:2503.19786},
  year    = {2025},
  url     = {https://arxiv.org/abs/2503.19786}
}

@article{qwen25_technical_report,
  title   = {Qwen2.5 Technical Report},
  author  = {{Qwen Team} and Yang, An and Yang, Baosong and Zhang, Beichen and Hui, Binyuan and Zheng, Bo and Yu, Bowen and Li, Chengyuan and Liu, Dayiheng and Huang, Fei and Wei, Haoran and Lin, Huan and Yang, Jian and Tu, Jianhong and Zhang, Jianwei and Yang, Jianxin and Yang, Jiaxi and Zhou, Jingren and Lin, Junyang and Dang, Kai and Lu, Keming and Bao, Keqin and Yang, Kexin and Yu, Le and Li, Mei and Xue, Mingfeng and Zhang, Pei and Zhu, Qin and Men, Rui and Lin, Runji and Li, Tianhao and Tang, Tianyi and Xia, Tingyu and Ren, Xingzhang and Ren, Xuancheng and Fan, Yang and Su, Yang and Zhang, Yichang and Wan, Yu and Liu, Yuqiong and Cui, Zeyu and Zhang, Zhenru and Qiu, Zihan},
  journal = {arXiv preprint arXiv:2412.15115},
  year    = {2025},
  url     = {https://arxiv.org/abs/2412.15115}
}

@misc{meta_llama33_70b_instruct,
  title        = {Llama 3.3 | Model Cards and Prompt Formats},
  author       = {{Meta}},
  howpublished = {\url{https://www.llama.com/docs/model-cards-and-prompt-formats/llama3_3/}},
  note         = {Model card for Llama 3.3 70B Instruct, accessed 2026-04-05}
}

@article{qwen3_technical_report,
  title   = {Qwen3 Technical Report},
  author  = {{Qwen Team}},
  journal = {arXiv preprint arXiv:2505.09388},
  year    = {2025},
  url     = {https://arxiv.org/abs/2505.09388}
}

@article{gptoss_model_card,
  title   = {gpt-oss-120b \& gpt-oss-20b Model Card},
  author  = {{OpenAI}},
  journal = {arXiv preprint arXiv:2508.10925},
  year    = {2025},
  url     = {https://arxiv.org/abs/2508.10925}
}

@misc{austin2021programsynthesislargelanguage,
      title={Program Synthesis with Large Language Models}, 
      author={Jacob Austin and Augustus Odena and Maxwell Nye and Maarten Bosma and Henryk Michalewski and David Dohan and Ellen Jiang and Carrie Cai and Michael Terry and Quoc Le and Charles Sutton},
      year={2021},
      eprint={2108.07732},
      archivePrefix={arXiv},
      primaryClass={cs.PL},
      url={https://arxiv.org/abs/2108.07732}, 
}

@misc{sander2024watermarkingmakeslanguagemodels,
      title={Watermarking Makes Language Models Radioactive}, 
      author={Tom Sander and Pierre Fernandez and Alain Durmus and Matthijs Douze and Teddy Furon},
      year={2024},
      eprint={2402.14904},
      archivePrefix={arXiv},
      primaryClass={cs.CR},
      url={https://arxiv.org/abs/2402.14904}, 
}

@misc{pan2025llmwatermarksrobustlyprevent,
      title={Can LLM Watermarks Robustly Prevent Unauthorized Knowledge Distillation?}, 
      author={Leyi Pan and Aiwei Liu and Shiyu Huang and Yijian Lu and Xuming Hu and Lijie Wen and Irwin King and Philip S. Yu},
      year={2025},
      eprint={2502.11598},
      archivePrefix={arXiv},
      primaryClass={cs.CL},
      url={https://arxiv.org/abs/2502.11598}, 
}

@misc{xie2025trainingdataprovenanceverification,
      title={Training Data Provenance Verification: Did Your Model Use Synthetic Data from My Generative Model for Training?}, 
      author={Yuechen Xie and Jie Song and Huiqiong Wang and Mingli Song},
      year={2025},
      eprint={2503.09122},
      archivePrefix={arXiv},
      primaryClass={cs.CV},
      url={https://arxiv.org/abs/2503.09122}, 
}

@misc{zhang2024trainingdataattributionmodel,
      title={Training Data Attribution: Was Your Model Secretly Trained On Data Created By Mine?}, 
      author={Likun Zhang and Hao Wu and Lingcui Zhang and Fengyuan Xu and Jin Cao and Fenghua Li and Ben Niu},
      year={2024},
      eprint={2409.15781},
      archivePrefix={arXiv},
      primaryClass={cs.CV},
      url={https://arxiv.org/abs/2409.15781}, 
}

@misc{zhang2025detectingdistillationdatareasoning,
      title={Detecting Distillation Data from Reasoning Models}, 
      author={Hengxiang Zhang and Hyeong Kyu Choi and Sharon Li and Hongxin Wei},
      year={2025},
      eprint={2510.04850},
      archivePrefix={arXiv},
      primaryClass={cs.CL},
      url={https://arxiv.org/abs/2510.04850}, 
}

@misc{watson2022importancedifficultycalibrationmembership,
      title={On the Importance of Difficulty Calibration in Membership Inference Attacks}, 
      author={Lauren Watson and Chuan Guo and Graham Cormode and Alex Sablayrolles},
      year={2022},
      eprint={2111.08440},
      archivePrefix={arXiv},
      primaryClass={cs.CR},
      url={https://arxiv.org/abs/2111.08440}, 
}

@misc{adhikesaven2026modelsmodelsbuilton,
      title={Which Models Are Our Models Built On? Auditing Invisible Dependencies in Modern LLMs}, 
      author={Sanjay Adhikesaven and Haoxiang Sun and Sewon Min},
      year={2026},
      eprint={2606.12385},
      archivePrefix={arXiv},
      primaryClass={cs.CL},
      url={https://arxiv.org/abs/2606.12385}, 
}

\appendix
\clearpage

\section{Experimental Details}\label{app:exp_details}
\tightparagraph{Base student models ($S_0$)}
We consider Llama-3.2-3B-Instruct~\citep{meta_llama32_3b_instruct},
Gemma-3-4B-PT~\citep{gemma3_technical_report}, and
Qwen2.5-1.5B and Qwen2.5-3B~\citep{qwen25_technical_report}
as base student models.

\tightparagraph{Teacher models ($\mathcal{T}$)}
Our teacher candidate set includes
Llama-3.3-70B-Instruct~\citep{meta_llama33_70b_instruct},\footnote{
    We use the quantized version from \texttt{nvidia/Llama-3.3-70B-Instruct-NVFP8}.
}
Qwen3-8B~\citep{qwen3_technical_report},
Gemma-3-27B~\citep{gemma3_technical_report}, and
GPT-OSS-120B~\citep{gptoss_model_card}. %

These four models together form the pool of candidate teachers for the teacher identification task, allowing each to act as both a positive and a negative candidate.
Consequently, the task is formulated as a four-way classification task.
To prevent our task from becoming trivial due to substantial variation in teacher output lengths, we standardize teacher outputs prior to distillation by truncating each output to at most 2,048 tokens. 

\tightparagraph{Inputs ($\mathcal{X}$)} We use two open-source prompt datasets: OpenMathInstruct-2 (OMI-2), a dataset of math problems and reasoning questions, and s1, another set of reasoning-heavy prompts used for test-time scaling \citep{toshniwal2024openmathinstruct2,muennighoff2025s1}.  %

We create distilled student models by querying each teacher with input prompts, and fine-tuning each student on the resulting teacher-generated data. In total, this setup produces 4 base student models $\times$ 3 teacher models $\times$ 2 prompt datasets, yielding 24 distilled models. We then retain models only when distillation leads to performance improvements, outperforming their corresponding base models,\footnote{
    We use GSM8K and MATH 500 to evaluate whether distillation improves performance. Detailed results are reported in Table~\ref{tab:appendix_training_results}..
} resulting in 19 models. 

\paragraph{Baseline detection methods} We consider the following baselines. (1) \textbf{Raw likelihood}, which averages the token-level log-likelihood of $T_k(x_i)$ under $S$. (2) \textbf{ZLIB-normalized likelihood}, which normalizes likelihood using compression-based length estimates. (3) \textbf{Min-$k$\% probability}, which focuses on the lowest-probability tokens in the sequence. (4) \textbf{Similarity-based scoring}, which compares teacher and student outputs using lexical overlap (e.g., $n$-gram Jaccard) and embedding-based similarity. Finally, we compare to our primary method, \textbf{reference-based MIA}, which normalizes likelihood using a reference model. We also evaluate a \textbf{reverse MIA} variant, where we instead generate outputs from the student $S(x_i)$ and score them under each candidate teacher model $T_k$. Results are shown in Table~\ref{tab:teacher_identification_controlled_reverse}.

\paragraph{Evaluation Metrics}\label{appx:eval-metrics}
We evaluate teacher identification using two complementary decision rules. First, we report a sorted per-sample score: for each candidate teacher, we sort its alignment scores over the probe set and compare candidates at matched ranks, measuring how often the true teacher obtains the highest alignment score (Per-sample). Formally, letting $\mathbf{f}^{\downarrow}_{T_j,r}$ denote the $r$-th largest alignment score for teacher $T_j$ over $\hat{\mathcal{X}}$:
\[
\hat{T}_{\mathrm{sorted}}
=
\arg\max_{T_k \in \mathcal{T}}
\frac{1}{N}\sum_{r=1}^N
\mathbf{1}\!\left[
T_k
=
\arg\max_{T_j \in \mathcal{T}}
\mathbf{f}^{\downarrow}_{T_j,r}
\right].
\]
This captures whether the true teacher is favored distributionally, rather than requiring it to win on the same prompt. Second, we report aggregate identification accuracy by applying both decision rules at the model level: selecting the teacher with the highest mean alignment score,
\[
\hat{T}_{\mathrm{mean}}
=
\arg\max_{T_k \in \mathcal{T}}
\frac{1}{N}\sum_{i=1}^N \mathbf{f}(T_k(x_i), S),
\]
and selecting the teacher that wins most often under the sorted per-sample comparison. We then report the higher accuracy between these two aggregate decision rules (Agg.).

We report the models used in the main paper for our controlled experiments (Table~\ref{tab:appendix_training_results}), including only settings where distillation improves utility over the base model on both GSM8K and MATH500. When using our highly customized prompt template (OMI COT), we include all distilled models, observing gains in at least one of GSM8K or MATH500.

\tightparagraph{Distillation hyperparameters}\label{appx:distillation-hyperparameters}
For each (student, teacher, prompt dataset) combination, we fine-tune the base student $S_0$ on the teacher-generated dataset $\mathcal{D}_T$ using supervised fine-tuning. We use identical hyperparameters across all controlled distillation runs: 3 epochs, learning rate $1\mathrm{e}{-5}$ with cosine schedule and 5\% warmup, per-device batch size 4 with gradient accumulation 4 (effective batch size 16), block size 4{,}096, bf16 precision, and gradient checkpointing. Loss is computed on response tokens only; prompt tokens are masked with $-100$.

For Llama-3.2-3B-Instruct, we apply the model's native chat template (\texttt{apply\_chat\_template}) to format each (question, response) pair, and identify the prompt span via the prompt-only tokenization with \texttt{add\_generation\_prompt=True} so that supervision is restricted to the assistant turn. For the non-instruct base students (Gemma-3-4B-PT, Qwen-2.5-1.5B, Qwen-2.5-3B), we use a simple \texttt{Problem:\textbackslash n\{question\}\textbackslash n\textbackslash nSolution:\textbackslash n} template. Teacher responses are truncated to 2{,}048 tokens prior to training to prevent variation in teacher output length making our detection task easier, and the same truncation is applied when computing reference-based MIA scores at evaluation time. 

\tightparagraph{Compute resources}\label{appx:compute-resources}
We ran all training and evaluation on H200 GPUs. Each controlled distillation run uses 2$\times$H200, as do all reference-based MIA evaluations and baselines (similarity-based, logit-based, reverse MIA, ablations, and the o1 ASCII-vs-Unicode diagnostic). Larger candidate teachers were run on the same 2$\times$H200 setup, except for DeepSeek-R1, which we ran locally on 8$\times$H200 due to its size. Outputs from API-only models were generated through their respective APIs and did not require local GPUs.

We additionally report descriptive statistics for teacher output lengths in Table~\ref{tab:avg_teacher_output_lengths}. These statistics provide context for our controlled distillation setup, since teacher models differ substantially in output verbosity, especially when full reasoning traces are included. For models with explicit reasoning traces, we also report final-output-only lengths to separate reasoning verbosity from the final answer length.

\begin{table}[t]
\centering
\resizebox{\textwidth}{!}{%
\begin{tabular}{l l l l c c c c}
\toprule
ID & Student & Teacher & Data & Template & Trace & GSM8K & MATH500 \\
\midrule
B1 & Qwen-2.5-1.5B (Base) & Baseline & -- & -- & -- & 67.25 & 32.6 \\
   & Qwen-2.5-1.5B & Nvidia-Llama-3.3-70B-Instruct & OMI-2 (1K) & None & No & 68.68 & 34.0 \\
   & Qwen-2.5-1.5B & Nvidia-Llama-3.3-70B-Instruct & s1 (1K) & None & No & 69.44 & 34.0 \\
   & Qwen-2.5-1.5B & Nvidia-Llama-3.3-70B-Instruct & OMI-2 (918) & OMI COT & No & 69.00 & 38.0 \\
   & Qwen-2.5-1.5B & Qwen-3-8B & OMI-2 (1K) & None & Yes & 68.40 & 37.4 \\
   & Qwen-2.5-1.5B & Qwen-3-8B & s1 (1K) & None & Yes & 69.75 & 36.8 \\
   & Qwen-2.5-1.5B & GPT-OSS-120B & OMI-2 (1K) & None & Yes & 69.59 & 39.8 \\
   & Qwen-2.5-1.5B & GPT-OSS-120B & s1 (1K) & None & Yes & 69.67 & 40.6 \\
\midrule
B2 & Qwen-2.5-3B (Base) & Baseline & -- & -- & -- & 75.82 & 39.8 \\
   & Qwen-2.5-3B & Nvidia-Llama-3.3-70B-Instruct & OMI-2 (1K) & None & No & 76.87 & 43.2 \\
   & Qwen-2.5-3B & Nvidia-Llama-3.3-70B-Instruct & s1 (1K) & None & No & 76.40 & 42.6 \\
   & Qwen-2.5-3B & Nvidia-Llama-3.3-70B-Instruct & OMI-2 (918) & OMI COT & No & 77.00 & 47.6 \\
   & Qwen-2.5-3B & Qwen-3-8B & OMI-2 (1K) & None & Yes & 79.98 & 47.6 \\
   & Qwen-2.5-3B & Qwen-3-8B & s1 (1K) & None & Yes & 79.37 & 45.6 \\
   & Qwen-2.5-3B & GPT-OSS-120B & OMI-2 (1K) & None & Yes & 79.15 & 51.0 \\
   & Qwen-2.5-3B & GPT-OSS-120B & s1 (1K) & None & Yes & 79.90 & 53.4 \\
\midrule
B3 & Llama-3.2-3B-Instruct (Base) & Baseline & -- & -- & -- & 76.57 & 42.0 \\
   & Llama-3.2-3B-Instruct & Nvidia-Llama-3.3-70B-Instruct & OMI-2 (1K) & None & No & 77.25 & 54.2 \\
   & Llama-3.2-3B-Instruct & Nvidia-Llama-3.3-70B-Instruct & OMI-2 (918) & OMI COT & No & 77.71 & 40.8 \\
   & Llama-3.2-3B-Instruct & Nvidia-Llama-3.3-70B-Instruct & s1 (1K) & None & No & 76.88 & 48.0 \\
\midrule
B4 & Gemma-3-4B-pt (Base) & Baseline & -- & -- & -- & 38.13 & 23.6 \\
   & Gemma-3-4B-pt & GPT-OSS-120B & s1 & None & Yes & 53.37 & 29.4 \\
   & Gemma-3-4B-pt & GPT-OSS-120B & OMI-2 (1K) & None & Yes & 51.17 & 27.6 \\
   & Gemma-3-4B-pt & Nvidia-Llama-3.3-70B-Instruct & s1 & None & No & 50.57 & 27.4 \\
   & Gemma-3-4B-pt & Nvidia-Llama-3.3-70B-Instruct & OMI-2 (918) & OMI COT & No & 50.27 & 23.0 \\
   & Gemma-3-4B-pt & Qwen-3-8B & s1 & None & Yes & 51.78 & 28.05 \\
   & Gemma-3-4B-pt & Qwen-3-8B & OMI-2 (1K) & None & Yes & 48.59 & 28.2 \\
\bottomrule
\end{tabular}%
}
\caption{Student model training results for the controlled distillation setups.}
\label{tab:appendix_training_results}
\end{table}

\begin{table}[t]
\centering
\small
\makebox[\textwidth][c]{%
\begin{tabular}{lrr}
\toprule
Model & Avg Words (with s1 prompts) & Avg Words (with OMI prompts) \\
\midrule
Llama-3.3-70B-Instruct (Final output only) & 524.6 & 402.7 \\
Gemma-3-27B-it (Final output only) & 565.5 & 419.5 \\
Claude-3.5-Sonnet (Final output only) & 289.4 & 229.7 \\
Claude Opus 4.5 & 2776.7 & 1838 \\
Claude Opus 4.6 & 1646 & 1398.1 \\
GPT-OSS-120B & 1800.1 & 1032.5 \\
o1 & 840.2 & 687.1 \\
o3 & 658.6 & 758.9 \\
DeepSeek R1 & 5085.7 & 3577.6 \\
DeepSeek R1 (Final output only) & 284.1 & 225.0 \\
Qwen-3-8B & 8369.9 & 5708.1 \\
Qwen-3-8B (Final output only) & 517.0 & 385.3 \\
QwQ-32B Preview & 4701.0 & 3301.0 \\
\bottomrule
\end{tabular}%
}
\caption{\textbf{Average output length in words across teacher models.} We report the average number of words generated by each model under s1 prompts and OMI prompts.}
\label{tab:avg_teacher_output_lengths}
\end{table}

\section{Additional Experimental Results}\label{app:exp_results}
 \begin{table}[t]
\centering
\footnotesize
\setlength{\tabcolsep}{4pt}
\begin{tabular}{lrrrr}
\toprule
\multirow{2}{*}{\textbf{Reference model $R$}} & \multicolumn{2}{c}{\textbf{Llama 3.3-70B-Instruct as teacher}}  & \multicolumn{2}{c}{\textbf{GPT-OSS-120B as teacher}} \\
\cmidrule(lr){2-3} \cmidrule(lr){4-5}
& \textbf{Agg.} & \textbf{Per-sample} & \textbf{Agg.} & \textbf{Per-sample} \\
\midrule
    Llama 3.2-3B-Instruct (default) &  100.0 & 100.0  &  100.0 & 100.0 \\
    Llama 3.2-3B                    &  100.0 & 100.0  &    0.0 &  1.5 \\
    Llama 3.1-8B-Instruct           &  100.0 & 99.5  &  100.0 & 100.0 \\
    Llama 3.1-8B                    &  100.0 & 100.0 &    0.0 &  17.5 \\
\bottomrule
\end{tabular}
\caption{\textbf{Effect of the reference model $R$ on teacher identification}, when the base student model is Llama 3.2 3B Instruct. We note that using GPT-OSS-120B as a teacher did not lead to major utility, but we include it for diversity in this ablation.}
\label{tab:ref-model-ablation}
\end{table}

\begin{table}[t]
\centering
\footnotesize
\begin{tabular}{l rr rr rr}
\toprule
  \multirow{2}{*}{Method}
      & \multicolumn{2}{c}{Identical}
      & \multicolumn{2}{c}{Similar-domain}
      & \multicolumn{2}{c}{Different-domain} \\
  \cmidrule(lr){2-3} \cmidrule(lr){4-5} \cmidrule(lr){6-7}
  & Agg. & Per-sample
  & Agg. & Per-sample
  & Agg. & Per-sample
  \\
\midrule
  Random & 25.0 & 25.0 & 25.0 & 25.0 & 25.0 & 25.0 \\
\midrule
  \textbf{\em Length Matching Baseline}
      & 36.8 & 32.8
      & 36.8 & 34.9
      & 42.1 & 42.2 \\
\midrule
  \multicolumn{7}{l}{\textbf{\em String-based methods}} \\
  Multi n-gram Jaccard
      & 47.4 & 43.5
      & 47.4 & 41.1
      & 47.4 & 42.2 \\
  Symmetric Coverage
      & 47.4 & 46.4
      & 42.1 & 41.2
      & 52.6 & 44.3 \\
  Embedding Cosine
      & 42.1 & 41.8
      & 42.1 & 39.7
      & 36.8 & 30.8 \\
\midrule
  \multicolumn{7}{l}{\textbf{\em Logit-based methods}} \\
  Raw likelihood
      & 15.8 & 18.4
      & 10.5 & 14.6
      & 36.8 & 36.8 \\
  ZLIB-normalized
      & 36.8 & 40.3
      & 36.8 & 37.1
      & 36.8 & 36.8 \\
  Min-$k$
      & 21.1 & 19.9
      & 10.5 & 17.9
      & 36.8 & 36.8 \\
  Reverse raw likelihood
      & 47.4 & 43.7
      & 47.4 & 40.3
      & 52.6 & 45.3 \\
  Reverse ZLIB-normalized
      & 42.1 & 43.1
      & 42.1 & 41.7
      & 42.1 & 45.4 \\
  Reverse Min-$k$
      & 42.1 & 41.6
      & 36.8 & 39.1
      & 57.9 & 45.6 \\
\midrule
  \textbf{Ours} (Reference-based)
      & 100.0 & 100.0
      & 100.0 & 100.0
      & 100.0 & 98.8 \\
\bottomrule
\end{tabular}
\caption{
  \textbf{Impact of the choice of proxy input prompts ($\hat{\mathcal{X}}$)}.
  `Identical' indicates $\mathcal{X}$ and $\hat{\mathcal{X}}$ are identical;
  `Similar domain' indicates $\mathcal{X}$ and $\hat{\mathcal{X}}$ are not identical but are from similar domains; and
  `Different domain' indicates $\mathcal{X}$ and $\hat{\mathcal{X}}$ are from substantially different domains.
  Detection accuracy is relatively stable across all settings.
}
\label{tab:teacher_identification_proxy_input_ablation}
\end{table}

\subsection{Ablations on the choice of reference model ($R$)}\label{subsec:abl_ref_model}
We ablate the choice of reference model R to assess its impact on teacher identification (Table~\ref{tab:ref-model-ablation}). Using Llama 3.2-3B-Instruct students, we vary $R$ across related base and instruction-tuned models. Overall, performance is reasonably robust to the choice of $R$; however, using base (non-instruction-tuned) models degrades accuracy. This suggests that although the method is reasonably stable, matching the instruction-tuning of the student is important for reliable identification.

\subsection{Ablations on proximity in input prompts ($\hat{\mathcal{X}}$)}\label{subsec:abl_input_prompt_approximity}
To quantify the effect of input prompt approximation---i.e., the mismatch between the true distillation prompts $\mathcal{X}$ and the prompts that a detector has access to $\hat{\mathcal{X}}$---we consider three settings: \begin{enumerate}[leftmargin=14pt, topsep=1pt,itemsep=0pt]
    \item \textbf{Identical}: $\mathcal{X}$ and $\hat{\mathcal{X}}$ are identical, e.g., both are s1 prompts, or both are OMI prompts.
    \item \textbf{Similar-domain}: $\mathcal{X}$ and $\hat{\mathcal{X}}$ are not identical but are from similar domains, e.g., ($\mathcal{X}=$s1, $\hat{\mathcal{X}}=$OMI), or ($\mathcal{X}=$OMI, $\hat{\mathcal{X}}=$s1). This was the default setting in our main experiments.
    \item \textbf{Different-domain}: $\mathcal{X}$ and $\hat{\mathcal{X}}$ are from substantially different domains, e.g., $\mathcal{X}$ is s1 or OMI (math), while $\hat{\mathcal{X}}$ consists of prompts for code generation.\footnote{We took input prompts from MBPP \citep{austin2021programsynthesislargelanguage}}
\end{enumerate}
We follow the same setup as in the main experiments: models are distilled using four base students, four candidate teachers, and two prompt sets (s1 or OMI), and the detection task is to identify the true teacher among four candidates. Table~\ref{tab:teacher_identification_proxy_input_ablation} reports the results across these three proxy-input settings. Our method remains accurate even when the detector uses prompts from a different domain (code probing data), suggesting that the teacher-identification signal is not limited to the exact prompt distribution used during distillation.

\FloatBarrier
\subsection{Ablations for Reverse MIA methods}
\label{subsec:reverse-mia-method}

\begin{table}[H]
\centering
\scriptsize
\begin{tabular}{l rr rr rr}
\toprule
  \multirow{2}{*}{Method}
      & \multicolumn{2}{c}{Llama 3.3 70B Instruct}
      & \multicolumn{2}{c}{Qwen-3-8B}
      & \multicolumn{2}{c}{GPT-OSS-120B} \\
  \cmidrule(lr){2-3} \cmidrule(lr){4-5} \cmidrule(lr){6-7}
  & Agg. & Per-sample
  & Agg. & Per-sample
  & Agg. & Per-sample
  \\
\midrule
  Random
      & 25.0 & 25.0
      & 25.0 & 25.0
      & 25.0 & 25.0 \\
\midrule
  \multicolumn{7}{l}{\textbf{\em Reverse logit-based methods}} \\
  Reverse raw likelihood
      & 100.0 & 89.1
      & 33.3 & 23.7
      & 0.0 & 0.0 \\
  Reverse ZLIB-normalized
      & 100.0 & 93.7
      & 16.7 & 22.8
      & 0.0 & 0.1 \\
  Reverse Min-$k$
      & 100.0 & 97.2
      & 0.0 & 10.3
      & 0.0 & 0.2 \\
\midrule
  \textbf{Ours} (Reference-based)
      & 100.0 & 100.0
      & 100.0 & 100.0
      & 100.0 & 99.9 \\
\bottomrule
\end{tabular}
\caption{
    \textbf{Teacher identification accuracy (\%) for reverse logit-based baselines}.
    Accuracy is broken down by true teacher model, including both aggregated and per-sample metrics.
    Reverse likelihood variants perform well for Llama 3.3 70B Instruct but fail to reliably identify Qwen-3-8B and GPT-OSS-120B, while our reference-based method remains accurate across teachers.
}
\label{tab:teacher_identification_controlled_reverse}
\end{table}

\section{Detecting Distillation with a Highly Customized Prompt template}\label{app:appendix-OMI-2-COT-plots}
We provide additional visualization of the few-shot prompting strategy described in the main paper (Table~\ref{tab:teacher_identification_controlled_customized_instructions}). While the table reports aggregate detection accuracy, the plots below illustrate how few-shot prompting affects the distribution of alignment scores across candidate teachers.

Each subplot corresponds to a different distilled student model, and shows how candidate teachers behave when prompted with a small number of student-generated examples. We observe that few-shot prompting generally sharpens the separation between the true teacher and alternative candidates, improving identification performance. However, for the Gemma-3-4B-pt student, the separation remains weaker, consistent with the failure case observed in Table~\ref{tab:teacher_identification_controlled_customized_instructions}.

\begin{center}
    \includegraphics[height=0.19\textheight]{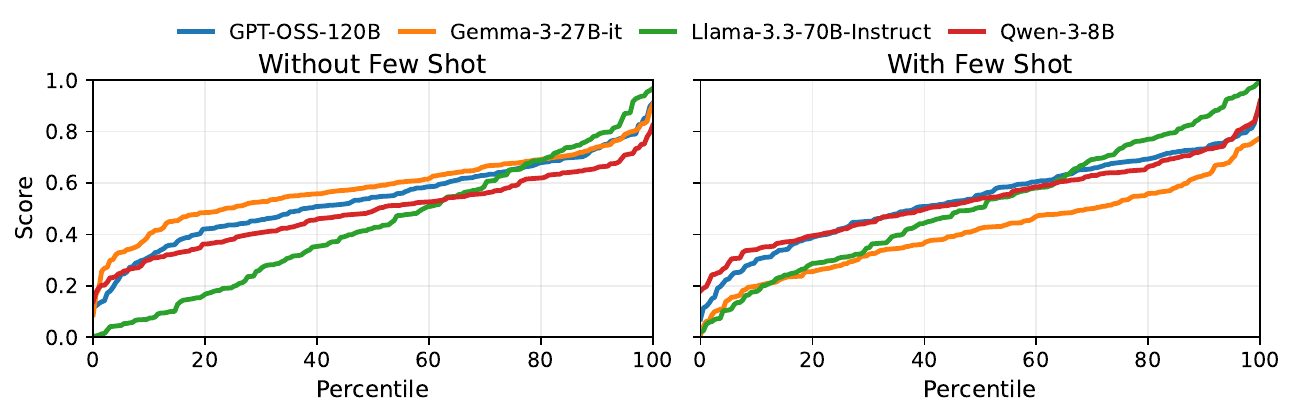}
    {\scriptsize (a) Qwen-2.5-1.5B\par}

    \includegraphics[height=0.19\textheight]{figures/OMI_Fewshot_Qwen-2.5-3B.pdf}

    {\scriptsize (b) Qwen-2.5-3B\par}

    \includegraphics[height=0.19\textheight]{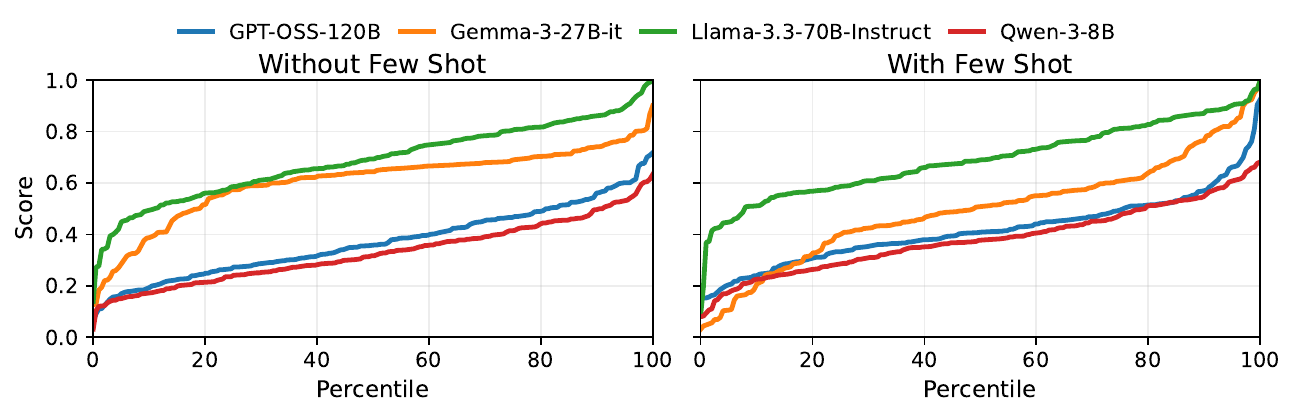}

    {\scriptsize (c) Llama-3.2-3B-Instruct\par}

    \includegraphics[height=0.19\textheight]{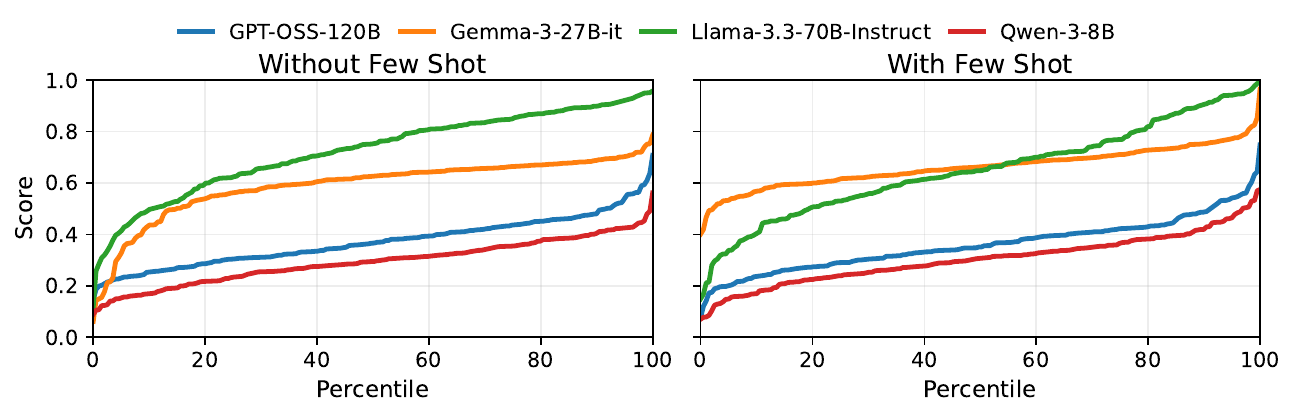}

    {\scriptsize (d) Gemma-3-4B-pt\par}

    \captionsetup{hypcap=false,font=small}
    \captionof{figure}{
    Results of prompting each candidate teacher model with few-shot examples drawn from the outputs of a distilled student model. Each subplot corresponds to a different student and evaluates how well candidate teachers can match the student's behavior under in-context learning. 
    }
    \label{fig:fewshot_teacher_matching}
\end{center}

\section{Model in the Wild Plots}\label{app:appendix-models-in-the-wild}
\begingroup
\raggedbottom

\captionsetup{
    hypcap=false,
    font=small,
    skip=2pt
}

\newcommand{\modelwildplot}[3]{%
    \par\noindent
    \begin{minipage}{\textwidth}
        \centering
        \makebox[\textwidth][c]{%
            \includegraphics[width=1.08\textwidth]{#1}%
        }
        \captionof{figure}{#2}
        \label{#3}
    \end{minipage}
    \par\vspace{0.5em}
}

We evaluate the DeepSeek-R1 distill models, \texttt{s1.1-32B}, and XCoder using reference-based membership inference (reference-based MIA). For each distilled model, we use as the reference model the corresponding base model from which the distill model was derived prior to distillation. We run the method on both OMI and s1 probe data, allowing us to test whether teacher-identification signals persist across related math-domain prompt distributions.

Overall, our method successfully identifies the true teacher across the evaluated candidates, showing that the reference-normalized likelihood signal remains informative even for real-world distilled models whose exact training data and distillation pipelines are unknown. This is encouraging because these models are not produced under our controlled setup: the prompts, filtering procedures, decoding parameters, and post-training details are all unobserved. Nevertheless, the true teacher generally receives the strongest alignment signal.

We further evaluate XCoder, a code-specialized model trained on code data, while still probing it using math data from s1 and OMI. The method continues to show a teacher-specific signal in this out-of-domain setting, suggesting that the attribution signal is not limited to models trained directly on math data.

\modelwildplot
    {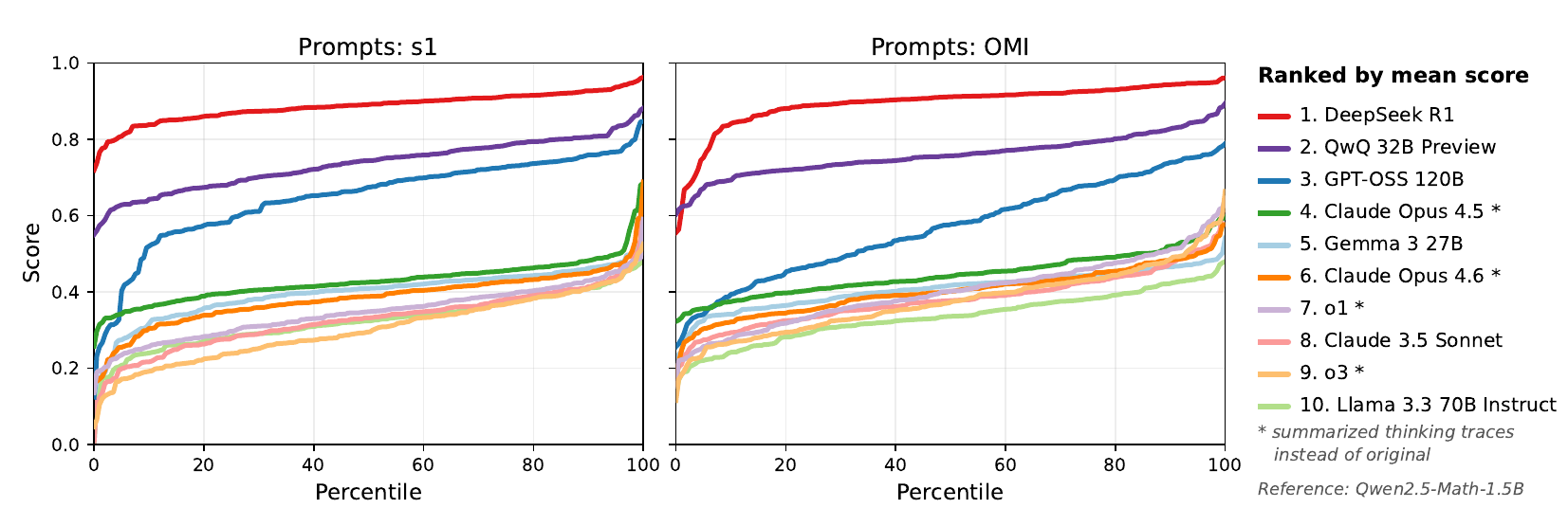}
    {Reference-based MIA results for \textbf{DeepSeek-R1-Distill-Qwen-1.5B}, using \textbf{Qwen2.5-Math-1.5B} as the reference model.}
    {fig:r1distill-qwen-1_5b}

\modelwildplot
    {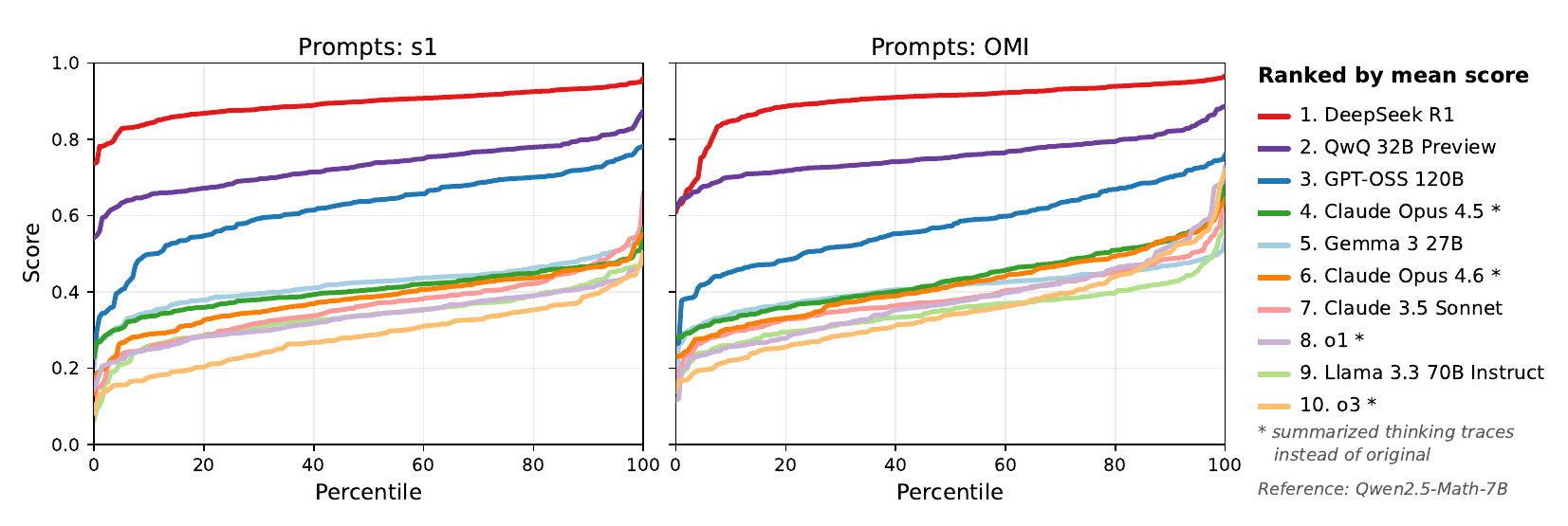}
    {Reference-based MIA results for \textbf{DeepSeek-R1-Distill-Qwen-7B}, using \textbf{Qwen2.5-Math-7B} as the reference model.}
    {fig:r1distill-qwen-7b}

\modelwildplot
    {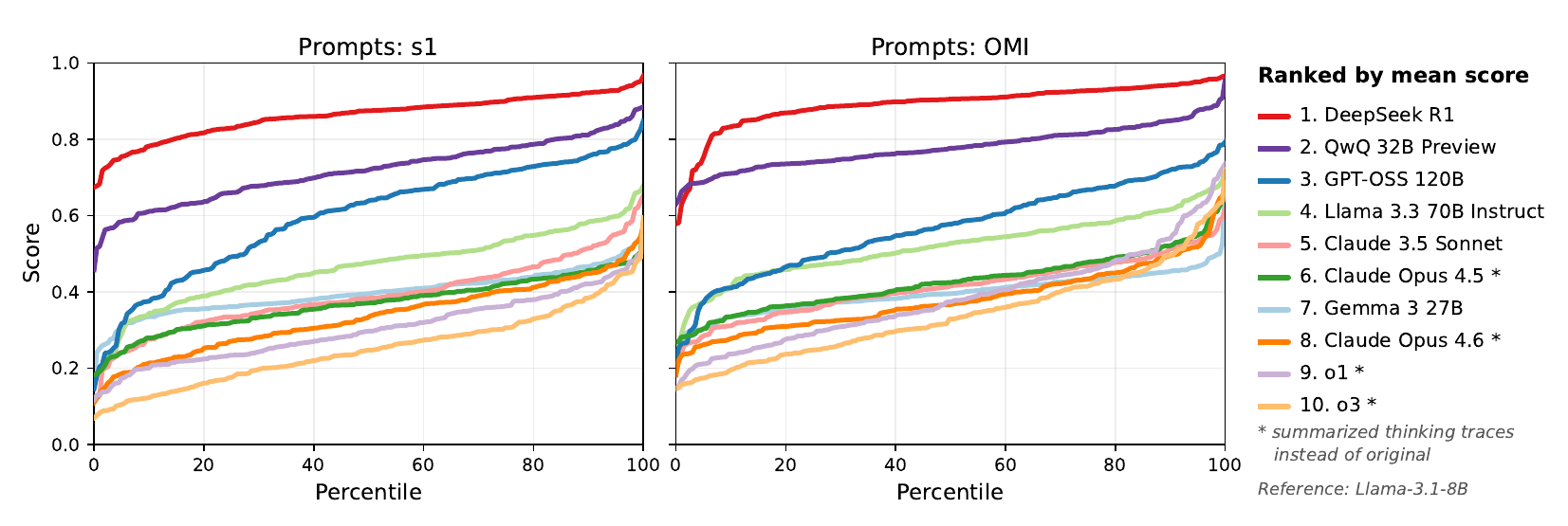}
    {Reference-based MIA results for \textbf{DeepSeek-R1-Distill-Llama-8B}, using \textbf{Llama-3.1-8B} as the reference model.}
    {fig:r1distill-llama-8b}

\modelwildplot
    {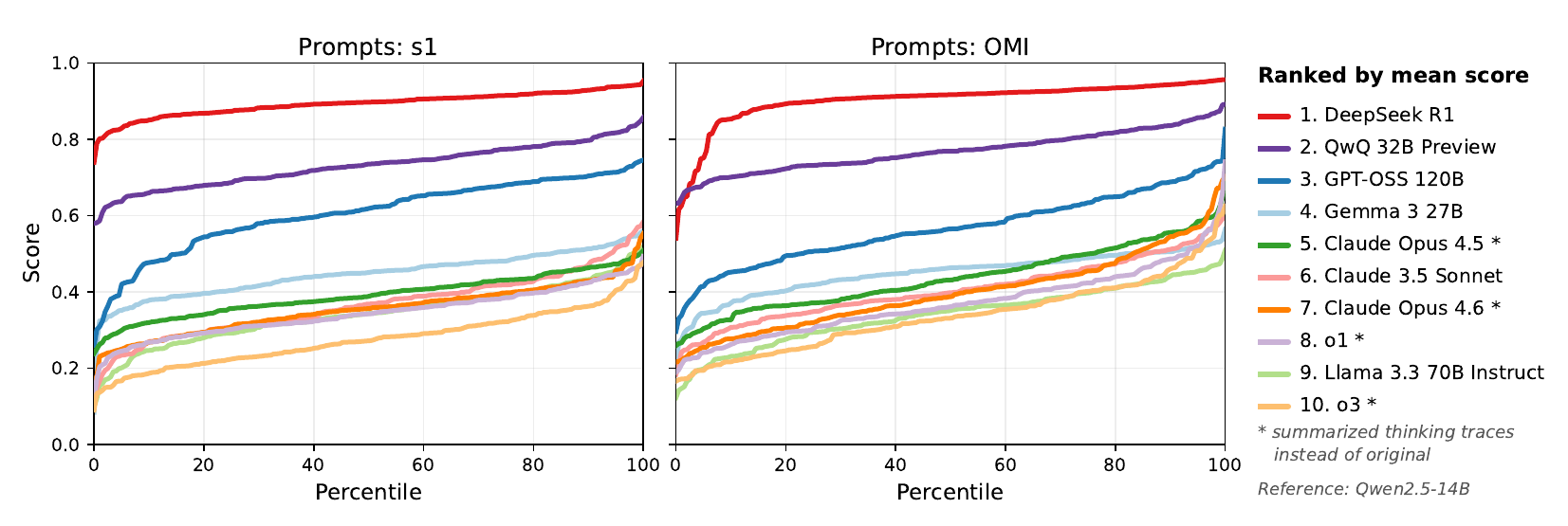}
    {Reference-based MIA results for \textbf{DeepSeek-R1-Distill-Qwen-14B}, using \textbf{Qwen2.5-14B} as the reference model.}
    {fig:r1distill-qwen-14b}

\modelwildplot
    {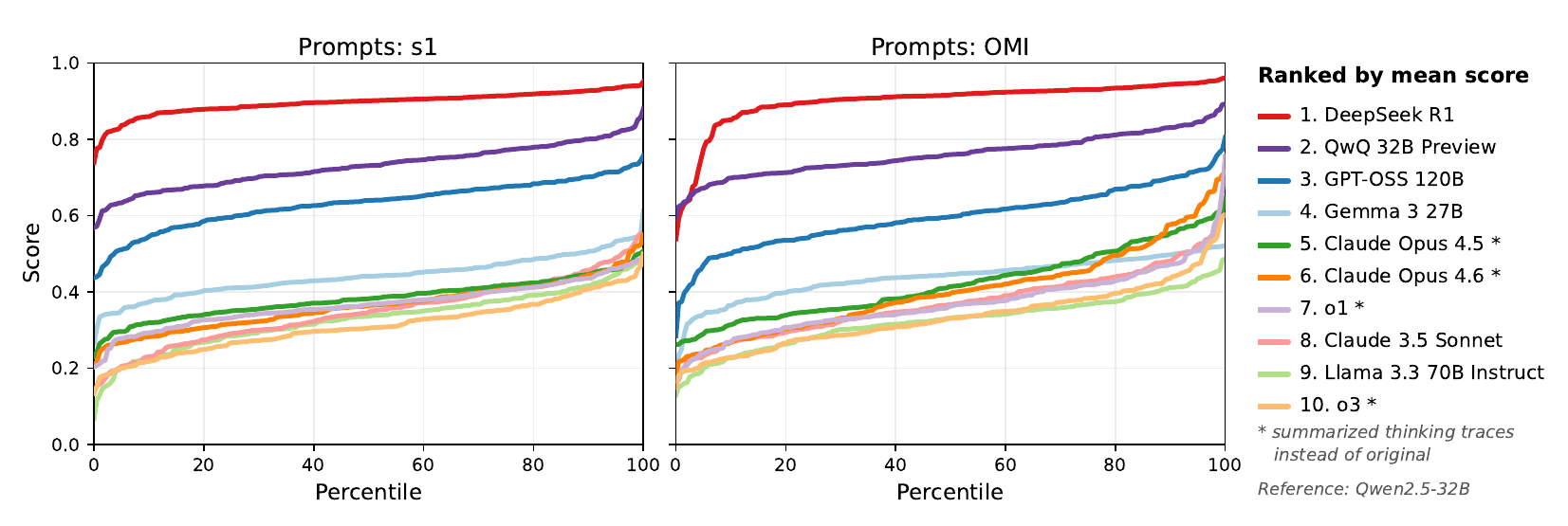}
    {Reference-based MIA results for \textbf{DeepSeek-R1-Distill-Qwen-32B}, using \textbf{Qwen2.5-32B} as the reference model.}
    {fig:r1distill-qwen-32b}

\modelwildplot
    {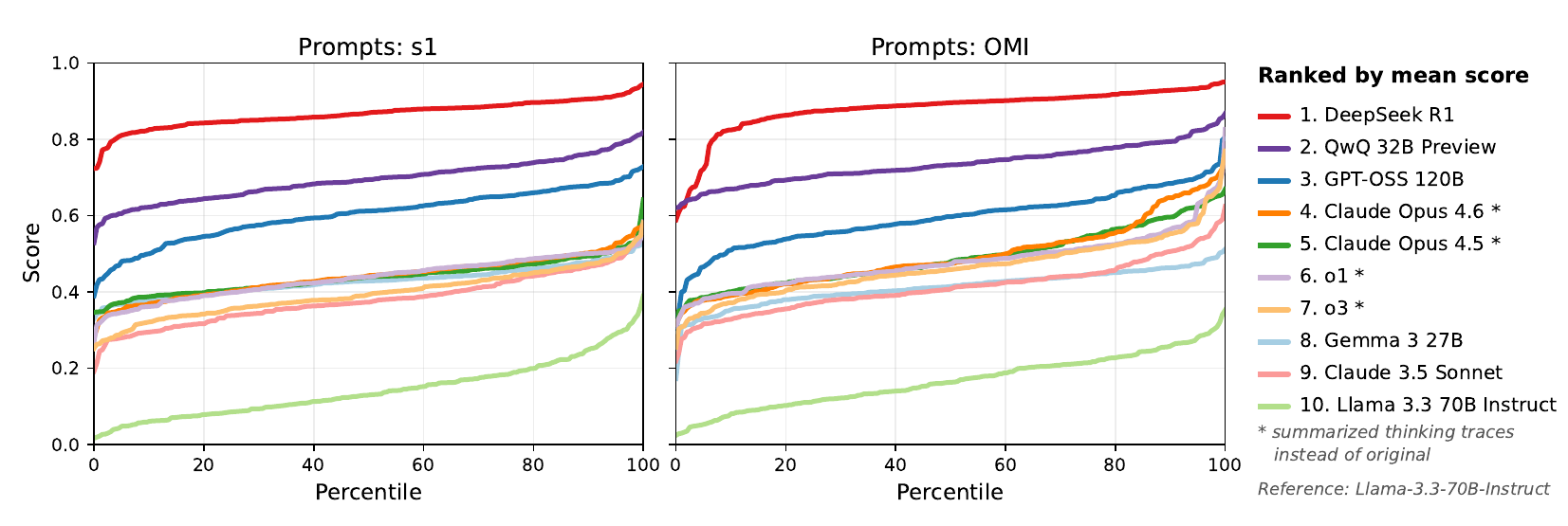}
    {Reference-based MIA results for \textbf{DeepSeek-R1-Distill-Llama-70B}, using \textbf{Llama-3.3-70B-Instruct} as the reference model.}
    {fig:r1distill-llama-70b}

\modelwildplot
    {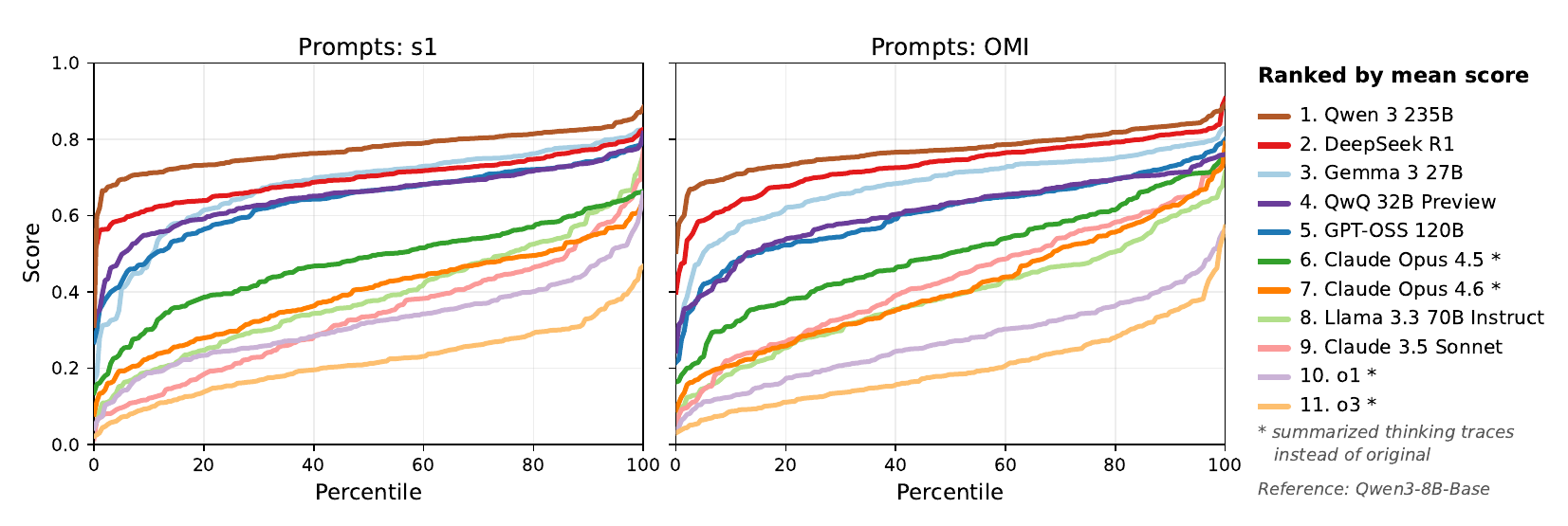}
    {Reference-based MIA results on \textbf{XCoder}. Although XCoder was trained on code data, we probe it using math datasets (s1 and OMI) to test whether teacher-specific likelihood signals remain detectable out of domain.}
    {fig:xcoder-ref}

\endgroup

\FloatBarrier
\section{Additional o1 Detection Details}\label{app:UnicodeResults}
\paragraph{Background: ASCII and Unicode.}
ASCII is a 7-bit encoding covering 128 characters; Unicode assigns codepoints to over 140{,}000 characters spanning nearly all writing systems and mathematical notation. ASCII is a strict subset of Unicode, so any Unicode string can be losslessly rendered as ASCII by escaping each non-ASCII codepoint as its literal \texttt{\textbackslash uXXXX} sequence.

\paragraph{Methodology.}
We generate each \texttt{o1} output once and serialize it two ways: a Unicode form (\texttt{ensure\_ascii=False}) preserving raw UTF-8 glyphs, and an ASCII form (\texttt{ensure\_ascii=True}) in which each non-ASCII codepoint becomes its six-character escape (e.g.\ ``\textbullet'' $\to$ \texttt{\textbackslash u2022}). For each target, we run reference-based MIA twice with all other hyperparameters fixed and define
$\Delta_{\text{ASCII}} = \overline{\mathcal{L}}_{\text{ASCII}} - \overline{\mathcal{L}}_{\text{Unicode}}$
on the reference-normalized loss. Under the null that the detector is not exploiting glyph-level features, $\Delta_{\text{ASCII}} \approx 0$; a positive gap indicates that ASCII normalization \emph{raises} the target's loss relative to the Unicode baseline---i.e., that the encoding contributes to the membership signal.

\paragraph{Controlled positives and probing data.}
To establish a ground-truth positive condition, we supervised-fine-tune four students---\texttt{Qwen2.5-1.5B}, \texttt{Qwen2.5-3B}, \texttt{Llama-3.2-3B-Instruct}, and \texttt{gemma-3-4b-pt}---with \texttt{o1} as the teacher on 1{,}000 questions drawn from the s1 dataset. Because the OpenAI API exposes only a \emph{summarized} reasoning trace, each training target is the concatenation of the summarized trace and the final answer. The MIA reference for each student is its own pre-SFT base checkpoint, isolating any signal acquired during distillation. As negative controls, we use real U.S.-organization models that are not expected to be distilled from \texttt{o1}/\texttt{o3}. We use 100 held-out OMI questions as probing data, providing a realistic setting where the detector does not have access to the original training inputs.

\paragraph{Additional control and wild-model results.}
We additionally evaluate a broader set of within-family controls, controlled \texttt{o1}-distilled positives, and wild models with large observed ASCII gaps. Table~\ref{tab:ascii_wild} reports the reference model used for each target and the resulting $\Delta_{\text{ASCII}}$.

\begin{table}[H]
\centering
\footnotesize
\begin{tabular}{l l r}
\toprule
Target & Reference & $\Delta_{\text{ASCII}}$ \\
\midrule
\multicolumn{3}{l}{\textit{Within-family controls (not distilled from \texttt{o1}/\texttt{o3})}} \\
Llama-3.1-8B-Instruct   & Llama-3-8B-Instruct       & $-0.0078$ \\
Llama-3.3-70B-Instruct  & Llama-3.1-70B-Instruct    & $+0.0007$ \\
Llama-3.1-8B            & Llama-3-8B                & $-0.0024$ \\
Llama-3.1-70B           & Llama-3-70B               & $-0.0050$ \\
gemma-2-9b-it           & gemma-1.1-7b-it           & $-0.0733$ \\
gemma-3-12b-it          & gemma-2-9b-it             & $+0.0373$ \\
gemma-3-27b-it          & gemma-2-27b-it            & $+0.0072$ \\
gemma-2-9b              & gemma-7b                  & $+0.0016$ \\
gemma-3-27b-pt          & gemma-2-27b               & $+0.0070$ \\
\midrule
\multicolumn{3}{l}{\textit{\texttt{o1}-distilled positives}} \\
gemma-3-4b-pt (SFT)        & gemma-3-4b-pt             & $+0.0732$ \\
Qwen2.5-1.5B (SFT)         & Qwen2.5-1.5B              & $+0.0594$ \\
Qwen2.5-3B (SFT)           & Qwen2.5-3B                & $+0.0573$ \\
Llama-3.2-3B-Instruct (SFT)& Llama-3.2-3B-Instruct     & $+0.0647$ \\
\midrule
\multicolumn{3}{l}{\textit{Models in the wild}} \\
DeepSeek-R1             & DeepSeek-MoE-16B-Base     & $+0.937$ \\
GPT-OSS-120B            & GPT-OSS-20B               & $+0.3207$ \\
GPT-OSS-120B            & GPT-2 XL                  & $+0.8310$ \\
GPT-OSS-20B             & GPT-2 XL                  & $+0.5103$ \\
\bottomrule
\end{tabular}
\caption{
\textbf{ASCII gap $\Delta_{\text{ASCII}}$ on controlled and wild models.}
For each target, we report the reference model used and $\Delta_{\text{ASCII}} = \overline{\mathcal{L}}_{\text{ASCII}} - \overline{\mathcal{L}}_{\text{Unicode}}$ on the reference-normalized loss, computed on 100 OMI probes (see \S\ref{sec:O1detection}). Within-family controls cluster near zero, controlled \texttt{o1}-distilled positives show consistently positive gaps, and \texttt{DeepSeek-R1} and both \texttt{GPT-OSS} models exhibit large positive gaps.
}
\label{tab:ascii_wild}
\end{table}

\begin{figure}[H]
    \centering
    \includegraphics[width=0.68\linewidth]{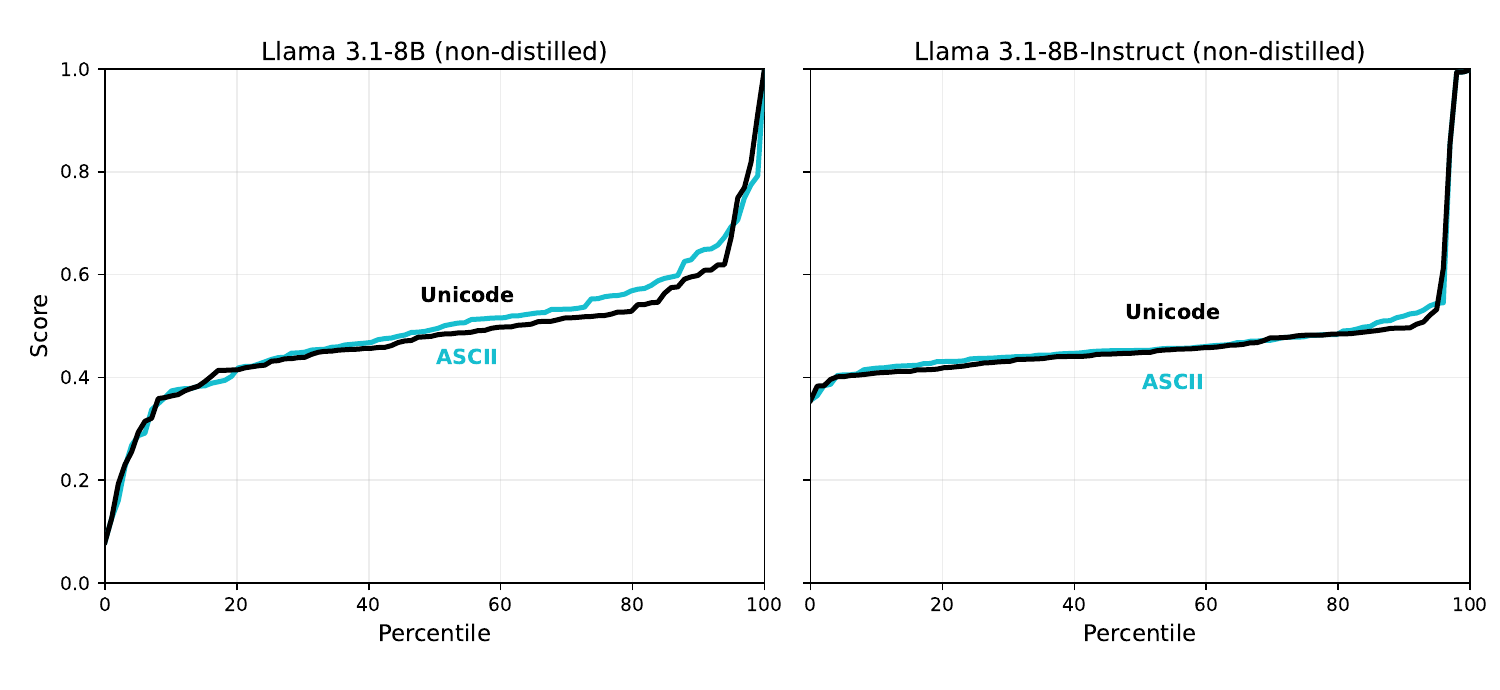}
    \caption{
    \textbf{ASCII vs.\ Unicode CDFs for non-\texttt{o1}-distilled Llama controls.}
    For Llama models not expected to be distilled from \texttt{o1}, the ASCII and Unicode reference-normalized loss distributions nearly overlap, indicating little to no encoding-specific separation.
    }
    \label{fig:llama_o1_nondistilled}
\end{figure}

The within-family controls---Llama and Gemma checkpoints not expected to be distilled from \texttt{o1} or \texttt{o3}---produce ASCII gaps concentrated near zero, with most pairs falling within $\pm 0.01$ and the largest deviation at $-0.0733$ for \texttt{gemma-2-9b-it}. This is also visible in Figure~\ref{fig:llama_o1_nondistilled}: for non-\texttt{o1}-distilled Llama controls, the ASCII and Unicode CDFs lie almost on top of each other, showing no substantial difference in reference-normalized loss under the two serializations. This is consistent with the hypothesis that no \texttt{o1}-specific encoding signal is present in models not exposed to \texttt{o1} during training, and matches the pattern reported in the main text.

In contrast, the controlled \texttt{o1}-distilled positives all exhibit positive gaps ($+0.0573$ to $+0.0732$), while the real-world models exhibit much larger positive gaps. \texttt{DeepSeek-R1}, with \texttt{DeepSeek-MoE-16B-Base} as reference, produces $\Delta_{\text{ASCII}} = +0.937$. Using \texttt{GPT-2 XL} as the reference, \texttt{GPT-OSS-20B} produces $\Delta_{\text{ASCII}} = +0.5103$ and \texttt{GPT-OSS-120B} produces $\Delta_{\text{ASCII}} = +0.8310$; using \texttt{GPT-OSS-20B} as a within-family reference for \texttt{GPT-OSS-120B} still yields $\Delta_{\text{ASCII}} = +0.3207$. Thus, the real-world gaps are not only positive but also substantially larger than those of our controlled \texttt{o1}-distilled students, often by nearly an order of magnitude. This mirrors the per-probe CDF separation for \texttt{GPT-OSS} and \texttt{DeepSeek-R1} reported in \S\ref{sec:openquestions}.

\tightparagraph{Surface-Level Stylistic Markers as Complementary Signals}
We additionally observe that certain teacher models exhibit highly consistent surface-level 
generation habits. Specifically, Llama-3.3-70B-Instruct responses follow a structured 
Markdown format using headings of the form \texttt{\#\# Step~$N$}, which 
\citet{toshniwal2024openmathinstruct2} document as the characteristic ``Llama CoT'' format. 
We similarly observe that \texttt{claude-3.5-sonnet} responses frequently contain the 
phrase ``Let's solve this step by step,'' and \texttt{QwQ-32B-Preview} responses frequently 
open with ``I've got this problem.'' Notably, \citet{toshniwal2024openmathinstruct2} show 
that a highly customized prompt template can suppress these stylistic artifacts during 
distillation, which may limit their utility as attribution signals. We leave a more 
systematic study---including robustness to prompt variation and integration with 
likelihood-based scores---to future work.

\tightparagraph{Results.}
Table~\ref{tab:surface_markers} reports occurrence rates across the first 200 OMI responses 
per teacher. The Llama marker is especially strong, appearing in every response with an 
average of 7.26 occurrences per response, while the Claude and QwQ markers occur at 
nontrivial but lower rates.

\begin{table}[h]
\centering
\small
\caption{Surface-level stylistic marker occurrence rates across the first 200 OMI responses 
per teacher model.}
\label{tab:surface_markers}
\begin{tabularx}{\linewidth}{lXccc}
\toprule
\textbf{Teacher Model} & \textbf{Marker} & \textbf{Responses Hit} & \textbf{Hit Rate (\%)} & \textbf{Avg.\ per Response} \\
\midrule
Claude 3.5 Sonnet      & ``Let's solve this step by step'' & 66 / 200  & 33.0  & 0.33 \\
QwQ-32B-Preview        & ``I've got this problem''          & 35 / 200  & 17.5  & 0.18 \\
Llama-3.3-70B-Instruct & \texttt{\#\# Step~$N$} heading    & 200 / 200 & 100.0 & 7.26 \\
\bottomrule
\end{tabularx}
\end{table}

\FloatBarrier

\section{Additional Open Question Plots}\label{app:WildModels}
This section provides additional reference-based MIA plots and statistical analysis for the open-question analyses in \S\ref{sec:openquestions}. These settings involve target models whose training data, prompts, instructions, filtering strategies, and distillation procedures are not publicly known, so the results should be interpreted as diagnostic rather than conclusive evidence of distillation.

\FloatBarrier

\begin{figure}[H]
    \centering
    \includegraphics[width=1.0\linewidth]{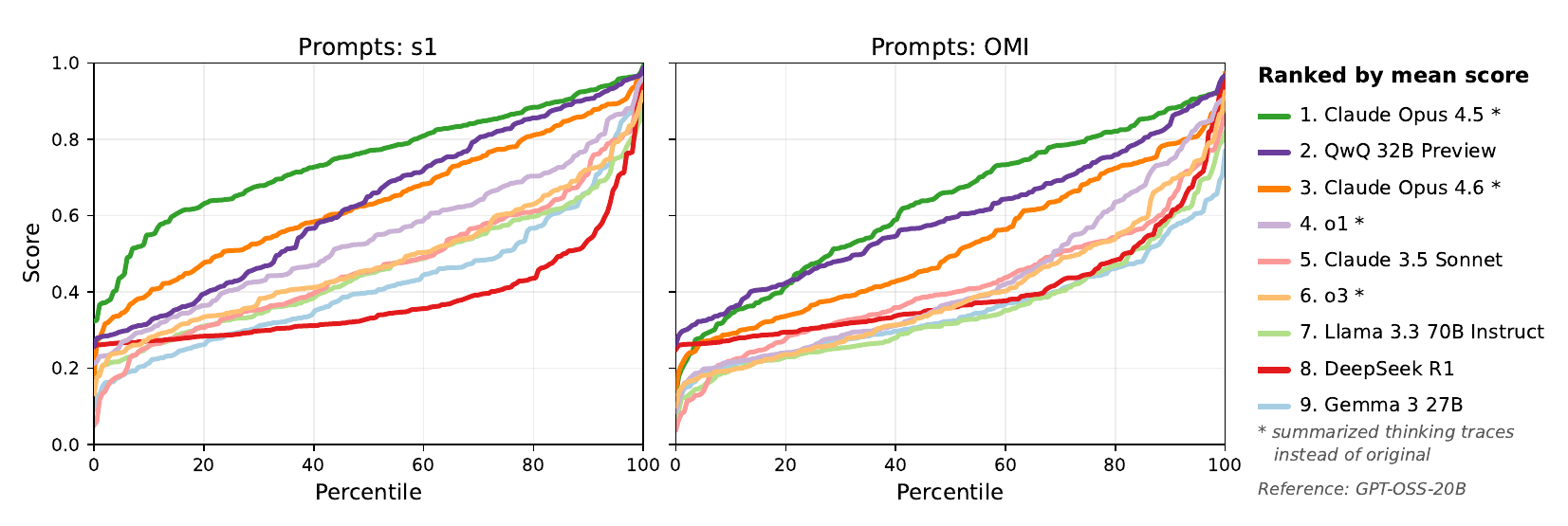}
    \vspace{-.7em}
    \caption{We apply reference-based MIA to \texttt{GPT-OSS-120B}, using \texttt{GPT-OSS-20B} as the reference model.}
    \label{fig:oss120b_oss20b}
\end{figure}

\begin{figure}[H]
\centering
\includegraphics[width=1.0\linewidth]{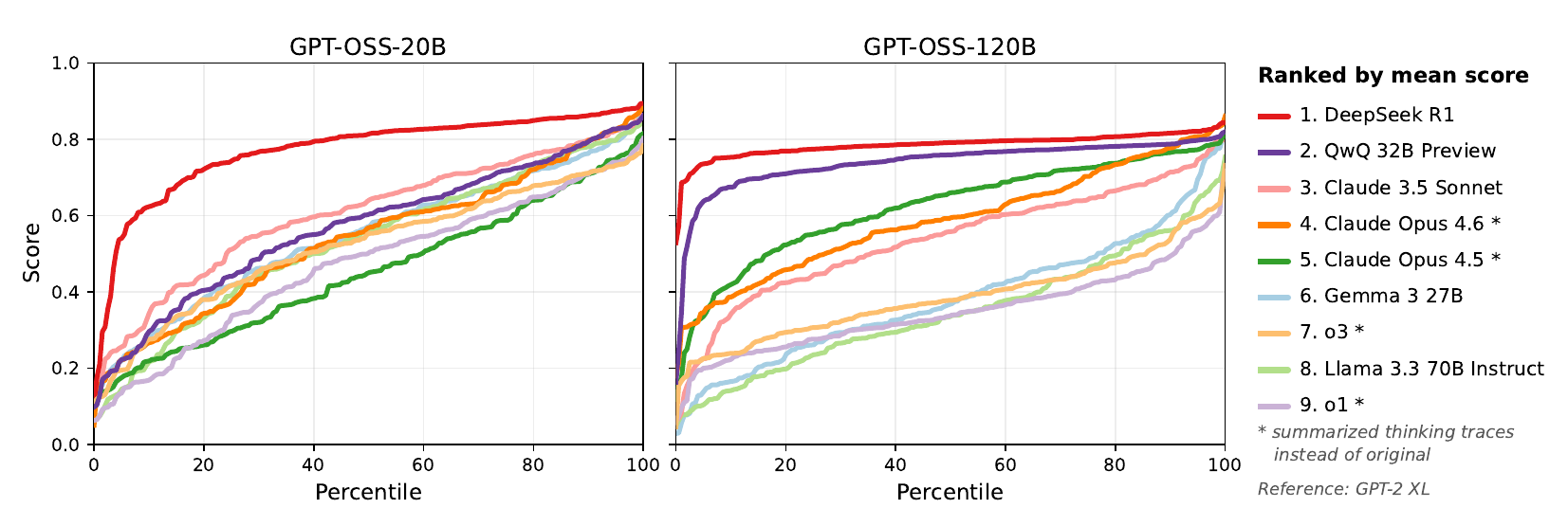}
\vspace{-.7em}
\caption{We apply reference-based MIA to \texttt{GPT-OSS-120B} and \texttt{GPT-OSS-20B}, using \texttt{GPT-2 XL} as the reference model and OMI as probing data.}
\label{fig:oss120b_2xl}
\end{figure}

\begin{figure}[H]
    \centering
    \includegraphics[width=1.0\linewidth]{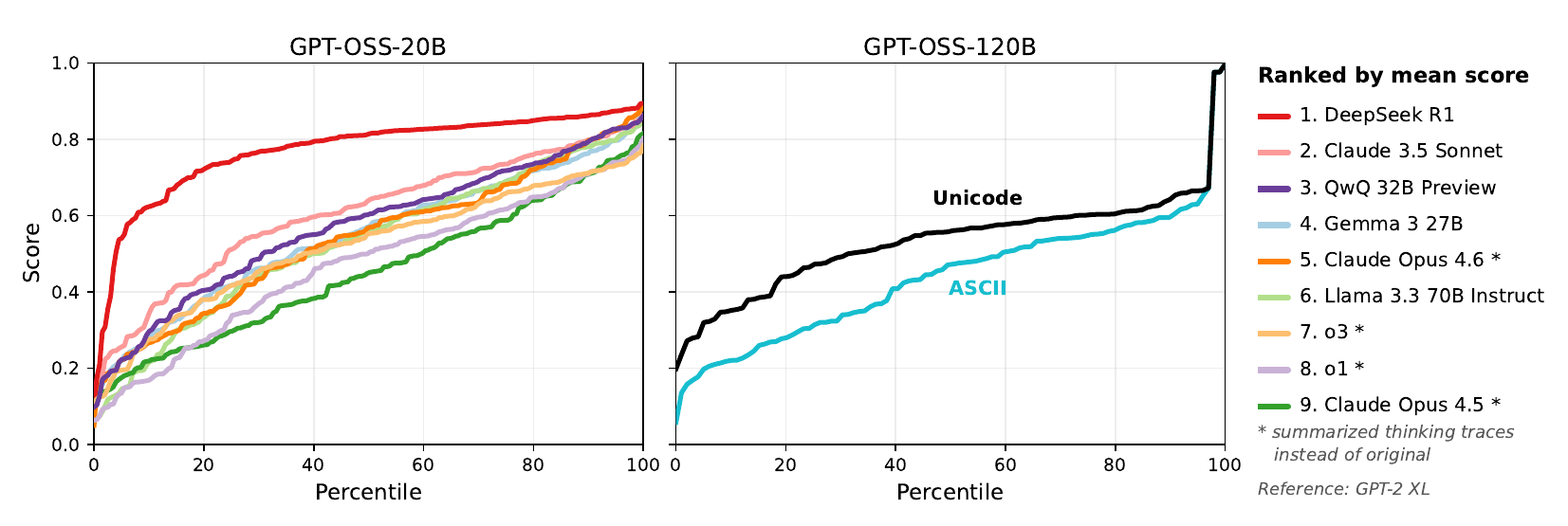}
    \vspace{-.7em}
    \caption{\textbf{Results for the GPT-OSS family.} We evaluate these models with both reference-based MIA (left) and our ASCII--Unicode diagnostic (right).}
    \label{fig:gptoss_combined_plot}
\end{figure}

\begin{table}[H]
\centering
\small
\setlength{\tabcolsep}{4pt}
\renewcommand{\arraystretch}{1.15}

\resizebox{\linewidth}{!}{%
\begin{tabular}{l l l r@{\hspace{3pt}}l l r c}
\toprule
Target & Reference & Probe &
\multicolumn{2}{c}{$\delta_{\text{Uni}-\text{ASCII}}$ [95\% CI]} &
Test & $p$ & Sig. \\
\midrule
\texttt{DeepSeek-R1}
& \texttt{DeepSeek-MoE-16B-Base}
& OMI
& $+0.937$
& \scriptsize$[+0.746,+1.141]$
& Wilcoxon
& $8.5{\times}10^{-16}$
& \cmark \\

\texttt{GPT-OSS-120B}
& \texttt{GPT-OSS-20B}
& OMI
& $+0.32$
& \scriptsize$[+0.21,+0.44]$
& Wilcoxon
& $3.5{\times}10^{-8}$
& \cmark \\

\texttt{GPT-OSS-120B}
& \texttt{GPT-2 XL}
& OMI
& $+0.83$
& \scriptsize$[+0.69,+0.98]$
& Wilcoxon
& $2.7{\times}10^{-16}$
& \cmark \\

\texttt{GPT-OSS-20B}
& \texttt{GPT-2 XL}
& OMI
& $+0.51$
& \scriptsize$[+0.39,+0.64]$
& Wilcoxon
& $1.3{\times}10^{-13}$
& \cmark \\
\bottomrule
\end{tabular}%
}
\caption{\textbf{Statistical significance of the \texttt{o1} Unicode--ASCII diagnostic.}
The within-model gap $\delta_{\text{Uni}-\text{ASCII}}$ tested one-sided against $0$ (Wilcoxon where
the differences are non-normal, $t$-test otherwise); \cmark~marks $p<0.05$. A significant gap means
the separation is resolvable rather than sampling noise---suggestive of \texttt{o1}-style signal,
not a confirmed distillation call.}
\label{tab:o1_sig}
\end{table}

\end{document}